\documentclass{article}

\usepackage{PRIMEarxiv}

\usepackage{amsmath}
\usepackage{algorithm}
\usepackage{algorithmic}

\usepackage[utf8]{inputenc}
\usepackage[T1]{fontenc}   
\usepackage{hyperref}      
\usepackage{url}           
\usepackage{booktabs}      
\usepackage{amsfonts}      
\usepackage{nicefrac}      
\usepackage{microtype}     
\usepackage{lipsum}
\usepackage{fancyhdr}      
\usepackage{graphicx}      
\graphicspath{{media/}}

\title{Personalized Federated Learning with Hidden Information on Personalized Prior}

\author{
    Mingjia Shi, Yuhao Zhou, Qing Ye, Jiancheng Lv 
    \thanks{The authors are with the College of Computer Science, Sichuan University, Chengdu 610065, China (e-mail: 310lihs@gmail.com; sooptq@gmail.com; yeqing@scu.edu.cn; lvjiancheng@scu.edu.cn).}
}

\begin{document}
\maketitle

\begin{abstract}
Federated learning (FL for simplification) is a distributed machine learning technique that utilizes global servers and collaborative clients to achieve privacy-preserving global model training without direct data sharing. However, heterogeneous data problem, as one of FL's main problems, makes it difficult for the global model to perform effectively on each client's local data. Thus, personalized federated learning (PFL for simplification) aims to improve the performance of the model on local data as much as possible. Bayesian learning, where the parameters of the model are seen as random variables with a prior assumption, is a feasible solution to the heterogeneous data problem due to the tendency that the more local data the model use, the more it focuses on the local data, otherwise focuses on the prior. When Bayesian learning is applied to PFL, the global model provides global knowledge as a prior to the local training process. In this paper, we employ Bayesian learning to model PFL by assuming a prior in the scaled exponential family, and therefore propose pFedBreD, a framework to solve the problem we model using Bregman divergence regularization. Empirically, our experiments show that, under the prior assumption of the spherical Gaussian and the first order strategy of mean selection, our proposal significantly outcompetes other PFL algorithms on multiple public benchmarks.
\end{abstract}


\section{Introduction}
\label{sec_intro}

In recent years, the growth of mobile devices, the proliferation of data volumes and data camping policies have inspired the development of federal learning (FL for simplification) \cite{mcmahan2017communication}, which mainly focuses on privacy preserving and communication. Even so, the highly heterogeneous data problem in federated settings are still exist \cite{tan2022towards}. Personalized federated learning (PFL) are motivated to alleviate this problem, aiming to find personalized models on local data in or after the federated training process.

One of the practical problems is that a single client has very little data with some certain labels, which means the local data distribution differs significantly from the global data distribution.\cite{kairouz2021advances,tan2022towards} However, local clients requires that the model perform better on the local distribution while also obtaining prior information (or knowledge) from the global data distribution from the global model. This makes it necessary to improve personalized performance and the generalization ability of the global model. The general method is to obtain a global model. The global model is directly used in local training as initialization or in the regularization term to penalize personalized model getting too far from global model, which requires to make a trade-off by the stepsize or regular terms, in the process of meta training or regularization-based training, as many PFL methods do \cite{tan2022towards}.

Briefly, different models are needed on different datasets, and these models require a priori information for training, which is passed through the distribution and aggregation of global models. This motivate us to model PFL with Bayesian learning and to personalize prior in each local optimization process making prior contain more local information. Main contributions of this paper are shown as follows:
\begin{itemize}
    \item We model a class of problem with Bayesian learning using personalized prior in the scaled exponential family.
    \item We propose a unified framework pFedBreD to solve this problem motivated by robust optimization, meta learning and mirror descent. The framework decouples MAML-like meta learning methods and Bregman divergence based regularization methods respectively as prior mean selection and regularization function selection.
    \item Empirically, we verify the performance of 3 personalized prior selection methods, under the prior assumption of the spherical Gaussian. Our method meets and exceeds most of the baselines on the public benchmarks.
\end{itemize}

\section{Related Works}
\label{sec_rltdw}

Federated learning is a distributed collaborative learning paradigm that preserves data privacy. Since the proposal of FedAvg\cite{mcmahan2017communication}, a growing number of challenges\cite{kairouz2021advances} in the federated learning setting (such as non-iid\cite{zhao2018federated, zhu2021federated}, client drift\cite{karimireddy2020mime, karimireddy2020scaffold}, and communication cost\cite{konevcny2016federated, zhou2021communication}, etc) have sparked heated debate. FedAvg attempts to lower communication costs while performing well on non-iid data, which is one of our test baselines. See \cite{kairouz2021advances} for additional information for federated learning.

Personalized approaches in the federated learning setting have been increasingly popular in recent years, and, as far as we know, most of them are based on regularization, or can be transformed into equivalent regularization-based methods. PFL methods can be roughly categorized as follows, see \cite{tan2022towards} for additional information of PFL:
\begin{itemize}
    \item \textbf{Regularization based Single Model:}
        Researchers have developed a variety of approaches based on regularization to handle the PFL challenge in recent years.(e.g. FedU\cite{dinh2021fedu}, pFedMe\cite{t2020personalized}, FedAMP\cite{huang2021personalized}, HeurFedAMP\cite{huang2021personalized})
        All of these approaches' personalized objective functions can be expressed as in Equation (\ref{equ_fml_owithr}):
        \begin{equation}
        \label{equ_fml_owithr}
            J(\theta) + R(\theta;\mu)
        \end{equation}
        where $J(\theta)$ is the loss function of the local problem and $R(\theta;\mu)$ is the regularization term used to restrict the deviation between $\theta$ and $\mu$.(e.g. $R(\theta;\mu)=\frac{1}{2}||\theta-\mu||^{2}$ in pFedMe)
    
    \item \textbf{Meta Learning based Single Model:}
        One of the most representative meta-learning based single-model PFL approach based is the well-known Per-FedAvg\cite{fallah2020personalized}, aiming to find an initialization that is easy to fine-tune. That is, the global model in the federated learning setting is regarded as a meta model in MAML\cite{fallah2020convergence}, where the objective function of the local problem is as in Formula (\ref{equ_fml_maml})
        \begin{equation}
        \label{equ_fml_maml}
            J(\theta-\eta\nabla J(\theta))
        \end{equation}
        Researchers also show the connections between FedAvg\cite{mcmahan2017communication} and Reptile\cite{nichol2018first}, another meta learning framework, in \cite{jiang2019improving}, and how to improve personalization of federated learning via Reptile. Proximal updating is also used in meta-learning based algorithms like \cite{zhou2019efficient}.
    
    \item \textbf{Multiple Model:}
        Federated Multi-task learning (FTML) and transfer learning have been proposed as several ways to improve the personalization of federated learning methods\cite{smith2017federated, wu2020personalized, li2019fedmd}.
        A big part of these approaches is based on knowledge distillation\cite{hinton2015distilling}, or mutual learning\cite{zhang2018deep}, a more flat multi-teacher version of knowledge distillation that is more efficient in distributed or federated learning realms\cite{anil2018large, jeong2018communication, shen2020federated}.
        A regularization with KL divergence approaches is proposed in \cite{yang2021h}, which uses run-time data augmentation during training. Public shareable datasets are utilized to distill knowledge between global and local models\cite{li2019fedmd}.
    
    \item \textbf{Expectation Maximization:}
        To our best knowledge, two Expectation Maximization \cite{dempster1977maximum} (EM for simplification) based methods are proposed, e.g. FedSparse in \cite{louizos2021expectation} and FedEM in \cite{dieuleveut2021federated}. Both of them focus on communication compression. The latter provides a variance reduce version and assumes complete information (or data) of the global model obeys distribution in exponential family. Another FedEM \cite{marfoq2021federated} combine FTML and EM.
    
    \item \textbf{Bayesian FL:}
        In recent years, studies of PFL using Bayesian optimization modeling have been proposed \cite{zhang2022personalized,achituve2021personalized,dai2020federated}. e.g. \cite{dai2020federated} uses Bayesian optimization via Thompson sampling.
        In related approaches, FLOA\cite{liu2021bayesian} and pFedGP\cite{achituve2021personalized} are proposed with KL-Divergence regularization in the loss function, which is comparable to applying specific a-prior assumptions in exponential family.
         Our implementation doesn’t use a Bayesian neural network (BNN) model as an inferential model as others do (e.g., pFedGP uses a Gaussian process tree and pFedBayes\cite{zhang2022personalized} uses BNN).

\end{itemize}

\section{Background Knowledge}

\subsection{Bregman-Moreau Envelope}
\label{ssec_bd}
Bregman divergence \cite{bregman1967relaxation} is defined in Equation (\ref{equ_def_bd}), where $g$ is a convex function. In this paper, for convenience, we discuss $g$ is strictly convex, proper and differentiable such that Bregman divergence is well-defined.
\begin{equation}
\label{equ_def_bd}
\begin{aligned}
\mathcal{D}_{g}(x,y) :&= g(x) - g(y) - \langle \nabla g(y) , x-y \rangle\\
& = \int_{y}^{x} \nabla g(t) - \nabla g(y) d t
\end{aligned}
\end{equation}
The definitions of Bregman proximal mapping and Bregman-Moreau envelope and the relationship between them are shown in Equation(\ref{equ_def_pmbme}) \cite{bauschke2006joint, bauschke2003bregman}, which we discuss as single-valued mappings and $\lambda > 0$.
\begin{equation}
\label{equ_def_pmbme}
\begin{aligned}
\mathcal{D}\mathbf{prox}_{g,\lambda^{-1}}f(x)&:=\arg\min_{\theta}\{f(\theta)+\lambda \mathcal{D}_{g}(\theta,x)\} \\
\mathcal{D}\mathbf{env}_{g,\lambda^{-1}}f(x)&:=\min_{\theta}\{f(\theta)+\lambda \mathcal{D}_{g}(\theta,x)\} \\
\nabla \mathcal{D}\mathbf{env}_{g,\lambda^{-1}}f(x)&=\lambda\nabla^{2}g(x)[x-\mathcal{D}\mathbf{prox}_{g,\lambda^{-1}}f(x)]
\end{aligned}
\end{equation}

\subsection{Exponential Family}

The regular exponential family \cite{banerjee2005clustering} (X-family for simplification) is chosen to yield our prior, which is shown in Equation (\ref{equ_def_ef}), where $g$ is assumed to be convex, $\mathcal{D}_{g}(\cdot,\cdot)$ is the Bregman divergence, and $g*$ is the Fenchel Conjugate of $g$.
\begin{equation}
\label{equ_def_ef}
\begin{aligned}
\mathbf{P}_{ef}(\mathcal{V};s,g)&=h(\mathcal{V})\exp\{\langle \mathcal{V}, s \rangle -g(s)\} \\
&=h(\mathcal{V})\exp\{-\mathcal{D}_{g^{*}}(\mathcal{V},\mu)+g^{*}(\mathcal{V})\},\mu=\nabla g(s)
\end{aligned}
\end{equation}
In Equation (\ref{equ_def_ef}), $s$, $h(\mathcal{V})$ and $g(s)$ are respectively the natural parameter, potential measure and logarithmic normalization factor. The complete information $\mathcal{V}$ is assumed to be the minimal sufficient statistic of it self.

The scaled exponential family is shown in Equation (\ref{equ_def_sef})
\begin{equation}
\label{equ_def_sef}
\begin{aligned}
\mathbf{P}_{sef}(\mathcal{V};\lambda,s,g)&=h_{\mathcal{V}}(\mathcal{V})\exp\{\lambda[\langle \mathcal{V}, s \rangle - g(s)]\} \\
&=h_{\lambda}(\mathcal{V})\exp\{-\lambda\mathcal{D}_{g^{*}}(\mathcal{V},\mu)+\lambda g^{*}(\mathcal{V})\}
\end{aligned}
\end{equation}

where $\log h_{\lambda}(\mathcal{V})$ is the scaled potential measure. The scaled X-family (SX-family) is still the X-family, whose scale parameter $\lambda$ is highlighted.

\section{Motivation}
\label{sec_mot}

 We use Bayesian modeling to define the main problem in federated learning with hidden information (HI), then EM-MAP to calculate (intuition from HI) and generate the expectation parameter of personalized prior via mirror descents’ iteration in each global epoch. In details, we discuss PFL modeled with Bayesian learning motivated by local hidden information in Section \ref{ssec_hi}, and select the prior in SX-family. The modeled main problem with selected prior is turned into bi-level robust optimization problem motivated by EM  in Section \ref{ssec_em}.\cite{ben2009robust} The personalized selection strategy of prior motivated by mirror descent and MAML is in Section \ref{ssec_spmmdl} and \ref{ssec_ppmaml}.

\subsection{Hidden Information}
\label{ssec_hi}

The general classification problem based on KL-divergence in the federated learning are formulated as Equation (\ref{equ_gfl})\footnote{More details of equations in this paper are in Appendix \ref{appdx_sec_doe}}.
\begin{equation}
\label{equ_gfl}
\begin{aligned}
&\arg\min_{w}\mathbf{E}_{i}\mathbf{E}_{d_{i}}\mathbf{KL}(\mathbf{P}(y_{i}|x_{i})||\hat{\mathbf{P}}(y_{i}|x_{i},w))\\
=&\arg\max_{w}\mathbf{E}_{i}\mathbf{E}_{d_{i}}\mathbf{E}_{y_{i}|x_{i}}\log\hat{\mathbf{P}}(y_{i}|x_{i},w),(x_{i},y_{i})\in d_{i}
\end{aligned}
\end{equation}
where we rewrite the discriminant model as a Maximum Likelihood Estimation (MLE) problem of $y_{i} | x_{i}$ in the last line of Equation (\ref{equ_gfl}). In deep learning, $(x_{i}, y_{i})$ represent the pairs of input and label respectively in dataset $d_{i}$ on the $i$-th client, and $\hat{\mathbf{P}}(y_{i}| x_{i},w)$ represents the output of the deep neural network for classification parameterized by $w$.

Because each particular local probability distribution is different, if we use the global model and local data as input during both inference and training, we are missing the hidden information that a particular data is sampled from a particular client. The particular information $\Theta_{i}$ on the $i$-th client is used to reformulate Equation (\ref{equ_gfl}) into Equation(\ref{equ_gfl_em}).
\begin{equation}
\label{equ_gfl_em}
\arg\max_{w}\mathbf{E}\log\int_{\Theta_{i}}\hat{\mathbf{P}}(y_{i}|x_{i},\Theta_{i},w)\mathbf{P}(\Theta_{i}|x_{i},w)d\Theta_{i}
\end{equation}
where $\mathbf{E} = \mathbf{E}_{i}\mathbf{E}_{d_{i}}\mathbf{E}_{y_{i}|x_{i}}$ for simplification.

\subsection{Expectation Maximization}
\label{ssec_em}

EM algorithm is used to compute a approximation of likelihood in unobserved variables (the hidden information in this paper) settings. We formulate MLE employing EM methods, as shown in Equation (\ref{equ_gfl_eml})\footnote{More details of equations are in Appendix \ref{appdx_ssec_em}}.
\begin{equation}
\label{equ_gfl_eml}
\begin{aligned}
& \sum_{i}\log\hat{\mathbf{P}}(y_{i}|x_{i},w) = \sum_{i}\log\int\hat{\mathbf{P}}(y_{i},\Theta_{i}|x_{i},w)d\Theta_{i} \\
=& \sum_{i}\log\int\mathbf{Q}(\Theta_{i})\frac{\hat{\mathbf{P}}(y_{i},\Theta_{i}|x_{i},w)}{\mathbf{Q}(\Theta_{i})}d\Theta_{i} \\
\ge& \sum_{i}\mathbf{E}_{\mathbf{Q}(\Theta_{i})}[\log\hat{\mathbf{P}}(y_{i}|x_{i},\Theta_{i},w)+\mathbf{E}_{y_{i}|x_{i},w}\log\mathbf{P}(\Theta_{i}|d_{i},w)
]
\end{aligned}
\end{equation}
where $\mathbf{Q}(\Theta_{i})$ is a probability measure of $\Theta_{i}$.

Assuming that prior $\Theta_{i}|d_{i},w \sim \mathbf{P}_{sef}(\Theta_{i};\lambda,s_{i}(w;d_{i}),g)$ and the local loss function on the $i$-th client $f_{i}(\Theta_{i},w)$ is $\mathbf{E}_{d_{i}}[-\log\mathbf{P}(y_{i}|x_{i},\Theta_{i},w)]$, we have the L.H.S in Equation (\ref{equ_gfl_sembme}) from (\ref{equ_gfl_eml}).

Here is a assumption for simplification that $\theta_{i}$ contains all the information for local inference, i.e. $\theta_{i} = \Theta_{i}$ and $\mathbf{P}(y_{i}|x_{i},\Theta_{i},w) = \mathbf{P}(y_{i}|x_{i},\Theta_{i})$.(e.g., $\theta_{i}$ has the same dimension of $w$ as the parameter of personalized model) Thus,  $f_{i}(\theta_{i})= \mathbf{E}_{d_{i}}[-\log\mathbf{P}(y_{i}|x_{i},\theta_{i})]$.

Thus, we can optimize a upper bound as a bi-level optimization problem as shown in Equation (\ref{equ_gfl_sembme})

in order to solve Equation (\ref{equ_gfl}) approximately, where mean parameter $\mu_{i}=\nabla g \circ s_{i}$. The following $d_{i}$ is omitted.($\mu_{i}(\cdot)\leftarrow\mu_{i}(\cdot;d_{i})$)

\begin{equation}
\label{equ_gfl_sembme}
-\max_{w,\{\theta_{i}\}}\mathbf{E}_{i}\{-f_{i}(\theta_{i})-\lambda\mathcal{D}_{g^{*}}(\theta_{i},\mu_{i}(w))\}
\le\min_{w}\mathbf{E}_{i}\min_{\{\theta_{i}\}}\{f_{i}(\theta_{i})+\lambda\mathcal{D}_{g^{*}}(\theta_{i},\mu_{i}(w))\} \\
\end{equation}

\subsection{Selecting Prior Mean via Mirror Descent}
\label{ssec_spmmdl}

At the very beginning of this section, we generally introduce a quotation from \cite{kunstner2021homeomorphic} to related basic knowledge:

\begin{quote}
    EM for the exponential family is a mirror descent algorithm.
\end{quote}

The mirror descent can be generally write as Equation (\ref{equ_def_md}) from the old $\hat{w}$ to the new one $\hat{w}^{+}$ in each iteration.\cite{mcmahan2017survey, kunstner2021homeomorphic}
\begin{equation}
\label{equ_def_md}
\hat{w}^{+}\leftarrow\arg\min_{\hat{\theta}}\{f(\hat{w})+\langle\nabla f(\hat{w}),\hat{\theta}-\hat{w}\rangle+\hat{\lambda}\mathcal{D}_{\hat{g}}(\hat{\theta},\hat{w})\}
\end{equation}
We rewrite the problem into a more general variant shown in Equation (\ref{equ_gfl_md}) with relaxed restrictions and superfluous parameter, assuming $\hat{\theta}$, $\hat{\mu}$ and $\hat{w}$ with the same dimension.
\begin{equation}
\label{equ_gfl_md}
\arg\min_{\hat{\theta},\hat{\mu}}\{\Psi(\hat{\theta}, \hat{w})+\langle\nabla\Phi(\hat{w}),\hat{\mu}-\hat{w}\rangle+\lambda\mathcal{D}_{g^{*}}(\hat{\theta},\hat{\mu})+(2\eta)^{-1}||\hat{\mu}-\hat{w}||^{2}\}
\end{equation}
We can transform Equation (\ref{equ_gfl_md}) back into Equation (\ref{equ_def_md}) by setting $\Phi(\hat{w})$ to make $\nabla\Phi(\hat{w})=\nabla f(\hat{w})$, and $\Psi(\hat{\theta},\hat{w})$ as a function with $f(\hat{w})$ and a penalty term to softly make $\hat{\theta}$ and $\hat{w}$ close as you want (e.g. $\hat{\lambda}\mathcal{D}_{\hat{g}}(\hat{\theta},\hat{w})$).

This provides us a way to extract $\mu$ the function to generate mean parameter of the prior we want as Equation (\ref{equ_gfl_mdlb}) in each local epoch, which is minimizing a upper bound of the problem in Equation (\ref{equ_gfl_md}).

\begin{equation}
\label{equ_gfl_mdlb}
\begin{aligned}
&\mathcal{D}\mathbf{env}_{g^{*},\lambda^{-1}}\Psi(\cdot,w)(\mu_{\Phi}(w))=\min_{\theta}\{\Psi(\theta,w)+\lambda\mathcal{D}_{g^{*}}(\theta,\mu_{\Phi}(w))\}\\
&\mu_{\Phi}(w)=\arg\min_{\mu}\{\langle\nabla\Phi(w),\mu-w\rangle + (2\eta)^{-1}||\mu-w||^{2}\}
\end{aligned}
\end{equation}

By optimality condition, we have $\mu_{\Phi}(w)= w - \eta\nabla\Phi(w)$, which can be specified by $\Phi$. The remaining part is actually a Bregman-Moreau envelope, and we can optimize the upper bound with an EM-MAP (maximum a-posterior) method, which is to calculate $\mu_{\Phi}(w)$ the mean and the proximal mapping $\mathcal{D}\mathbf{prox}_{g^{*},\lambda^{-1}}\Psi(\cdot,w)(\mu_{\Phi}(w))$ alternately.

\subsection{Personalized Prior and MAML}
\label{ssec_ppmaml}

We usually use some statistic to represent the distribution, and one of the applications is the reparameterization trick, when using Bayesian inference. Mean parameter is used to represent the prior under SX-family prior assumption given any $\lambda$ and $g$ in this paper.

Mean of the SX-family prior in Equation (\ref{equ_gfl_sembme}) is used in regularization term, which can be personalized in each client $i$ as $\mu_{i}$. We use $\mu_{i}(w)$ to generate prior in each iteration of our proposed framework pFedBreD through the process of online federated learning.

Motivated by this, we use pFedBreD with MAML to learn the personalized regularization (or personalized prior in Bayesian learning) in Section \ref{ssec_ime}. (e.g. $\mathbf{mfo}$ in Equation (\ref{equ_ime_ms}) is actually use MAML on the Bregman-Moreau envelope $\mathcal{D}\mathbf{env}_{g^{*},\lambda^{-1}}f_{i}$ instead of the local loss function $f_{i}$ by substituting it into $J$ in Equation (\ref{equ_fml_maml}).)

\section{Algorithm}
\label{sec_alg}

\subsection{Problem Formulation}
\label{ssec_probf}

In this section, we model PFL based on mentioned motivation in Section \ref{sec_mot}. Each personalized model $\theta_{i}$ and personalized mean of prior are the solution of $\mathcal{D}\mathbf{env}_{g^{*},\lambda^{-1}}f_{i}(\mu_{i}(w))$ and $\mu_{i}(w)$ on the $i$-th client, where $w$ is the global model.

Assuming local hidden information on the $i$-th client satisfies 
 $\theta_{i}|x_{i},w \sim \mathbf{P}_{sef}(\theta_{i};\lambda,s_{i}(w),g)$, the global optimization problem can be written as Equation (\ref{equ_main_problem_global}).
\begin{equation}
\label{equ_main_problem_global}
\begin{aligned}
    \min_{w}\mathbf{E}_{i}\{F_{i}(w):=\mathcal{D}\mathbf{env}_{g^{*},\lambda^{-1}}f_{i}(\mu_{i}(w))\}
\end{aligned}
\end{equation}
The given $g$ is strictly convex, $\lambda>0$, $f_{i}$ is the local loss function, $s_{i}(w)$ is the natural parameter and $\mu_{i}(w)=\mathbf{E}_{\theta_{i}|x_{i},w}\theta_{i}=\nabla g(s_{i}(w))$ is the expectation (or mean) parameter in Equation (\ref{equ_main_problem_global}).

\subsection{Framework: pFedBreD}
\label{ssec_framework}

In this section, we introduce a framework to solve the optimization problem we mentioned in Section \ref{ssec_probf}.

We use gradient based methods to solve the global problem using the gradient of $F_{i}$ shown in Equation (\ref{equ_gdmp}).
\begin{equation}
\label{equ_gdmp}
     \nabla F_{i}(w)=\lambda\mathbf{D}\mu_{i}(w)\nabla^{2}g^{*}(w)[w-\mathcal{D}\mathbf{prox}_{g^{*},\lambda^{-1}}f_{i}(\mu_{i}(w))]
\end{equation}
$\mathbf{D}$ represents the gradient operator of the vector value function, and $\nabla^{2}$ is the Hessian operator in Equation (\ref{equ_gdmp}).
There are three parts in Equation (\ref{equ_gdmp}) we need to deal with, and the first order methods are as shown below.

\subsubsection{Jacobian Matrix of Mean}
\label{sssec_alg_jmm}

Motivated by the prior selecting strategy mentioned in Section \ref{ssec_spmmdl}, we have $\mathbf{D}\mu_{i}(w)=I-\eta\nabla^{2} \Phi(w)$. Choosing different function $\Phi$ leads to different results. Here we use a first order method and get rid of the last term, i.e. $\mathbf{D}\mu_{i}\leftarrow I$, the identity function.

\subsubsection{Hessian Matrix}
\label{sssec_alg_hm}

To use the first order methods, we directly let $\nabla^{2}g^{*}(w)=I_{m}$  the identity matrix. It happens that assuming $\theta_{i}$ obeys the spherical Gaussian by letting $g=\frac{1}{2}||\cdot||^{2}$ leads to the same result. Moreover, we can assume $\theta_{i}$ obeys the general multivariable Gaussian by letting $g=\langle\cdot,\Sigma^{-1}\cdot\rangle$ and $\nabla^{2} g(w) = \Sigma^{-1} \succeq 0$.

\subsubsection{Proximal Mapping}
\label{sssec_alg_pm}

Given $\mu_{i}(w)$,  the proximal mapping part $\mathcal{D}\mathbf{prox}_{g^{*},\lambda^{-1}}f_{i}(\mu_{i}(w))$ can be approximately solved with numerical methods, e.g. gradient descent methods. In other words, we can alternately calculate $\mu_{i}(w)$ on each client and then fix $\mu_{i}(w)$ in each local epoch as mentioned in Section \ref{ssec_spmmdl}.

Thus, the framework to solve the main problem and generate each personalized model is shown in Algorithm \ref{alg_pFedBreD}, where $\mathcal{I}$ is the client selecting strategy for global model aggregation; $\mathcal{C}_{i,r}^{(t)}$ means the $i$-th local running context (The historically available and cache-able data entities on each client) at the $t$-th global iteration and the $r$-th local iteration; $w_{init}$ and $\theta_{init}$ are the selecting strategies on the $i$-th client; $\mu_{i}$ is the mean selecting strategy; $\alpha_{m}$ is the main problem stepsize; $\lambda$, $g$ are the chosen prior parameters and $T$, $R$, $N$ are respectively the number of total iterations, local iterations and clients. $\beta$ is used in the same trick as \cite{karimireddy2020scaffold, t2020personalized}.

\begin{algorithm}[tb]
\caption{pFedBreD framework}  
\label{alg_pFedBreD}
\textbf{Input}: $\mathcal{I}$,\{$d_{i}$\}, $i=1...N$\\
\textbf{Parameter}: $\alpha_{m}$,$g$,$\lambda$,$T$,$R$,\{$w_{init}$,$\theta_{i}$,$\mu_{i}$,\},$i=1...N$\\
\textbf{Output}: $w^{(T)}$,\{$\theta_{i}^{(T)}$\}, $i=1...N$
\begin{algorithmic}[1]
\STATE Initialize $w^{(0)}$, $\{\theta_{i}^{(0)}\}$, $\{\mathcal{C}_{i,R}^{0}\}$;
\FOR{t=1...T}
\STATE Server sends $w^{(t-1)}$ to clients;
\FOR{i=1...N in parallel on each clients}
\STATE Initialize $w_{i,0}^{(t)}\leftarrow w_{init}(w^{(t-1)};\mathcal{C}_{i,R}^{(t-1)})$;
\STATE Initialize $\theta_{i,0}^{(t)}\leftarrow \theta_{init}(w^{(t-1)};\mathcal{C}_{i,R}^{(t-1)})$;
\FOR{r=1...R}
\STATE Generate $\mu_{i,r}^{(t)}\leftarrow \mu_{i}(w_{i,r-1}^{(t)};\mathcal{C}_{i,r}^{(t)})$;
\STATE $\theta_{i,r}^{(t)}\leftarrow\mathcal{D}\mathbf{prox}_{g^{*},\lambda^{-1}}f_{i}(\mu_{i,r}^{(t)})$;
\STATE $w_{i,r}^{(t)}\leftarrow w_{i,r-1}^{(t)}-\alpha_{m}\nabla F_{i}(w_{i,r-1}^{(t)})$;
\ENDFOR
\ENDFOR
\STATE Server collects $\{w_{i,R}^{(t)}\}$;
\STATE Server: $w^{(t)}\leftarrow(1-\beta) w^{(t-1)} + \beta \mathbf{E}_{\mathcal{I}} w_{i,R}^{(t)}$;
\ENDFOR
\STATE \textbf{return} $w^{T}$, \{$\theta_{i}^{T}$\}.
\end{algorithmic}
\end{algorithm}

\subsection{Implementation: Maximum Entropy}
\label{ssec_ime}

In this section, we propose a first order implementation based on maximum entropy rule \cite{friedman1971jaynes, jaynes1957information} for local prior. Besides, the non-maximum entropy rule approach is also worth considering, but we focus on maximum entropy prior in this section. See \cite{seidenfeld1979not, genest1986combining, kass1996selection, gelman2013philosophy} for additional information of non-maximum entropy assumption.

Practically, three main parts of the pFedBreD framework are needed to be implement in practice, which are shown as follows:
\begin{itemize}
    \item $g$, the function to derive logarithmic normalization factor, determines which type of prior is used.
    \item $\{s_{i}\}$ or $\{\mu_{i}\}$, the functions to derive the natural parameter and mean parameter for the personalized local prior, determines which particular prior is used given $g$.
    \item $\{w_{init}\}$ and $\{\theta_{i}\}$, the functions to derive the local initialization points of global model and personalized model respectively, requires local context $\mathcal{C}_{i,r}^{(t)}$ to derive.
\end{itemize}

The Gaussian distribution is the maximum entropy continuous distribution among SX-family given $g$, $\mu_{i}$ (the first order moment of $\theta_{i}$) and $\lambda$ (the parameter for optimization which, to some extent, determines the second moment of $\theta_{i}$ especially under the prior assumption of the spherical Gaussian distribution). Thus, we choose the scaled norm square $g=g^{*}=\frac{1}{2}||\cdot||^{2}$ to make the prior a spherical Gaussian.

Now, all we need to do is to choose a $\mu_{i}$, given $g$ as the scaled norm square and $\lambda$ as a hyper-parameter. That is to say that we don't need to calculate the natural parameter $s_{i}$, the one to assume the prior $\theta_{i}|x_{i},w \sim \mathbf{P}_{sef}(\theta_{i};\lambda,s_{i}(w),g)$. Because, there is a duality between both functions $s_{i}$ of the natural parameter and $\mu_{i}$ of the mean parameter with given $g$, that $s_{i}=\nabla g^{*} \circ \mu_{i}$ and $\mu_{i}=\nabla g \circ s_{i}$. The mean parameter $\mu_{i}$ in our implementation is selected by the methods mentioned in Section \ref{ssec_spmmdl}. Moreover, we have $\nabla g = \nabla g^{*}=I$ under the prior assumption of the spherical Gaussian, which means $\mu_{i}=s_{i}$. We can choose a different $\Phi_{i}$ as shown in Equation (\ref{equ_def_spmmdl})\footnote{A variant of $\mathbf{mg}$ in Appendix \ref{appdx_ssec_map}}, by $\mu_{i}(w) = w - \eta\nabla\Phi_{i}(w)$ the strategy mentioned in Section \ref{ssec_spmmdl}.
\begin{equation}
\label{equ_def_spmmdl}
    \Phi_{i}=\left\{
    \begin{aligned}
    & f_{i}, & \mathbf{fo}\\
    & F_{i}, & \mathbf{mfo}\\
    & f_{i}+F_{i}, & \mathbf{mg}\\
    \end{aligned}
    \right.
\end{equation}
where we use $\nabla F_{i}(w)=\lambda[w-\mathcal{D}\mathbf{prox}_{g^{*},\lambda^{-1}}f_{i}(\mu_{i}(w))]$ directly, which means that we let $\mathbf{D}\mu_{i}\leftarrow I$ and use $\nabla^{2} g^{*} = I_{m}$ as the first order methods. The three implementations of $\mu_{i}$, i.e. $\mathbf{fo}$, $\mathbf{mfo}$ and $\mathbf{mg}$, represent \textit{first order}, \textit{memorized first order} and \textit{memorized gradients} respectively.

\textit{Memorized first order} and \textit{memorized gradients} mean that we choose $\nabla F_{i}(w_{i,r-1}^{(t)})$ the gradient of Bregman-Moreau envelope as $\eta[w_{i,R}^{(t-1)} - \theta_{i,r-1}^{(t)}]$, where $\eta \ge 0$ is a stepsize-like hyper-parameter and each local client memorizes $w_{i,R}^{(t-1)}$, their own local part of the latest global model $w^{(t)}$ at last global epochs, instead of $ \lambda[w_{i,r-1}^{(t)} - \theta_{i,r-1}^{(t)}]$ in practice. We hold the opinion that the gradients of the methods $\mathbf{mfo}$ and $\mathbf{mg}$ are more stable and contain more information than plainly use current $w_{i,r-1}^{(t)}$ during gradient descent iteration. Moreover, the first order of $\nabla F_{i}$ leads to less computation of gradient calculation than $\nabla f_{i}$. Each methods of mean parameter selection in practice are shown in Equation (\ref{equ_ime_ms}).($\mathbf{fo}$, $\mathbf{mfo}$ and $\mathbf{mg}$ are respectively from top to bottom.)
\begin{equation}
\label{equ_ime_ms}
    \mu_{i,r}^{(t)} \leftarrow \left\{
    \begin{aligned}
        & w_{i,r-1}^{(t)} - \eta_{\alpha} \nabla f_{i}(w_{i,r-1}^{(t)}) \\
        & w_{i,r-1}^{(t)} - \eta (w_{i,R}^{(t-1)}-\theta_{i,r-1}^{(t)}) \\
        & w_{i,r-1}^{(t)} - \eta_{\alpha} \nabla f_{i}(w_{i,r-1}^{(t)}) - \eta (w_{i,R}^{(t-1)}-\theta_{i,r-1}^{(t)}) \\
    \end{aligned}
    \right.
\end{equation}
where $\eta_{\alpha}$ and $\eta$ are the stepsize-like parameters, and each line represents the methods of $\mathbf{fo}$, $\mathbf{mfo}$, $\mathbf{mg}$ from top to bottom in details.

Functions to derive initialization points of $w_{i}$ and $\theta_{i}$ at each local epoch on the $i$-th client are $w_{i,0}^{(t)}\leftarrow w^{(t-1)}$ and $\theta_{i,0}^{(t)}\leftarrow \theta_{i,R}^{(t-1)}$.
\section{Experiments}
\label{sec_exp}

In this section, we evaluate implementations of pFedBreD with the spherical Gaussian prior (norm square regularization) and the first order methods including $\mathbf{fo}$, $\mathbf{mfo}$ and $\mathbf{mg}$ proposed in Section \ref{ssec_ime}.

\subsection{Settings}
\label{ssec_exps}

\subsubsection{Datasets and Models}
\label{sssec_dm}

\begin{table}[t]
    \centering
    \resizebox{.94\textwidth}{!}{    
    \begin{tabular}{l|c|c|c|c|c}
        \hline
         \textbf{Dataset} & CIFAR-10 &  FEMNIST & FMNIST & MNIST & Sent140\\
        \hline
         \textbf{Model} & CNN & MCLR/DNN & MCLR/DNN & MCLR/DNN & LSTM\\
        \hline
     
    \end{tabular}
    }
    \caption{Datasets and models.}
    \label{tbl_tasks}
\end{table}

\begin{figure}
    \centering
    \includegraphics[width=1.6in,height=0.8in]{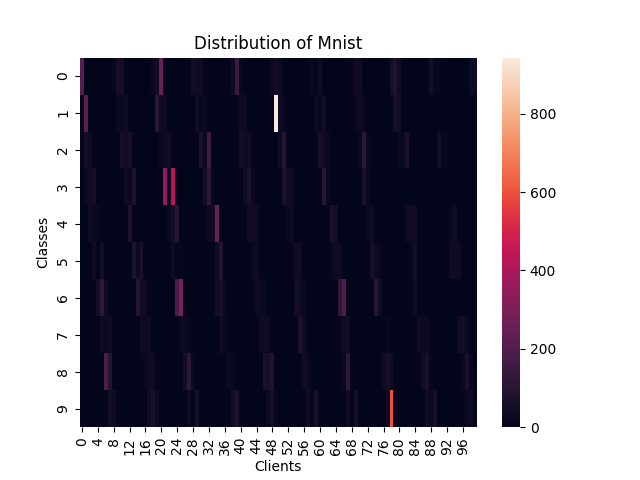}
    \includegraphics[width=1.6in,height=0.8in]{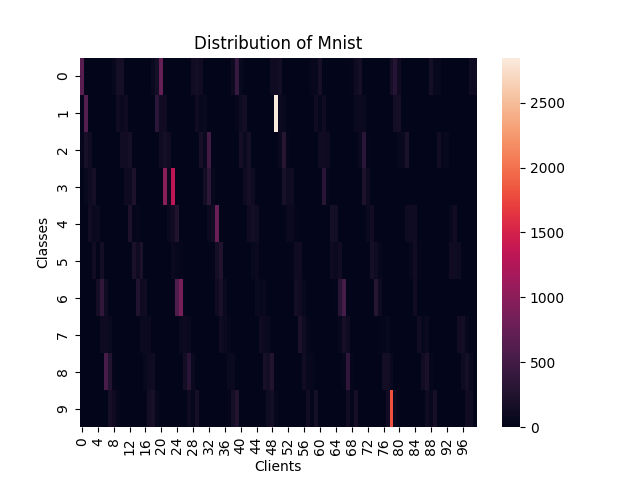}  
    \includegraphics[width=1.6in,height=0.8in]{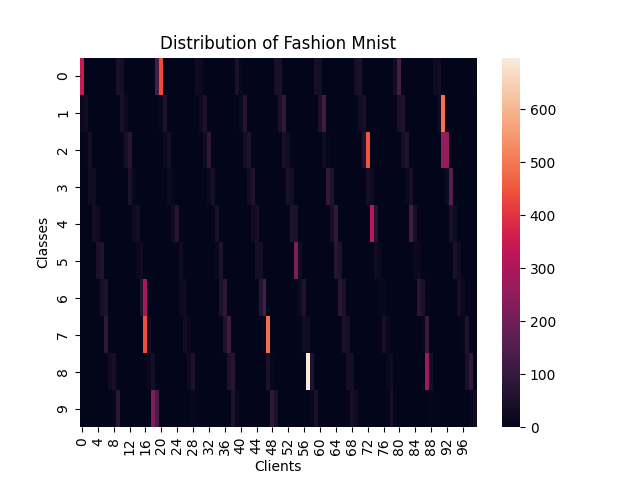} \includegraphics[width=1.6in,height=0.8in]{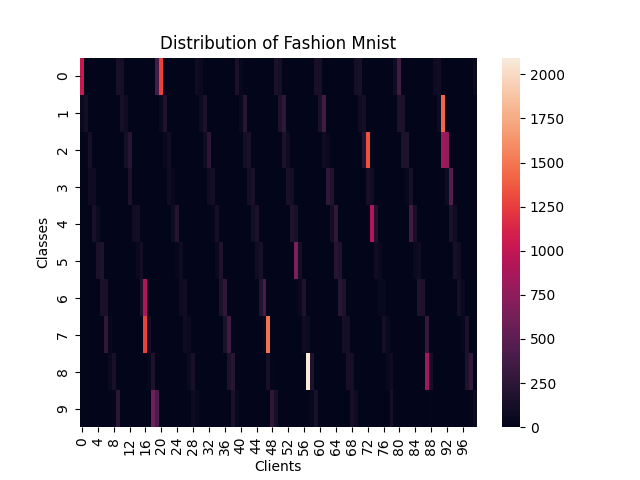}
    \begin{center}
        \footnotesize 
        MNIST-Test \qquad\qquad\qquad\qquad MNIST-Train \qquad\qquad\qquad\qquad
        FMNIST-Test \qquad\qquad\qquad\qquad FMNIST-Train
    \end{center}
    \caption{The visualization of the non-i.i.d. data distributions of MNIST and FMNIST.}
    \label{fig_niid}
\end{figure}

Our experiments\footnote{Access of all data and code is in Appendix \ref{appdx_sec_exp}} include several tasks as shown in Table \ref{tbl_tasks}. The non-i.i.d. settings of datasets are shown as follows and the non-i.i.d. distributions of MNIST and FMNIST are shown in Figure \ref{fig_niid} (more details are in Appendix \ref{appdx_sec_exp}):
\begin{itemize}
    \item \textbf{CIFAR-10}: The whole dataset is separated into 20 clients, and each client has data of 3 classes of label.\cite{dinh2021fedu, krizhevsky2009learning}
    \item \textbf{FEMNIST}: We use non-i.i.d. FEMNIST from LEAF benchmark with fraction of data to sample of 5\% and fraction of training data of 90\%.\cite{caldas2018leaf}
    \item \textbf{FMNIST}: The whole fashion-MNIST dataset is separated into 100 clients, and each client has data of 2 classes of label.\cite{t2020personalized, xiao2017fashion}
    \item \textbf{MNIST}: The whole MNIST dataset is separated into 100 clients, and each client has data of 3 classes of label.\cite{t2020personalized, lecun1998gradient}
    \item \textbf{Sent140}: We use non-i.i.d text dataset Sent140 from LEAF benchmark with fraction of data to sample of 5\%, fraction of training data of 90\% and minimum number of samples per user of 3. Then we re-separate Sent140 into 10 clients with at least 10 samples.\cite{caldas2018leaf}
\end{itemize}
Model settings are shown as follows:
\begin{itemize}
    \item \textbf{CNN:} For the image data, we use convolutional neural network of CifarNet \cite{hosang2015taking}.
    \item \textbf{DNN:} The non-linear model is 2 layers deep neural network with 100-dimension hidden layer and activation of leaky ReLU \cite{maas2013rectifier} and output of softmax.
    \item \textbf{MCLR:} The linear model, multi-class linear regression, is 1 layer of linear mapping with bias, and then output with softmax.
    \item \textbf{LSTM:} Text data model consists of 2 LSTM layers \cite{hochreiter1997long} as feature extraction layer of 50-dimension embeding and hidden layer and 2 layers deep neural network as classifier with 100-dimension of hidden layer.
\end{itemize}

\subsubsection{Baselines}
\label{sssec_bl}

We choose following algorithms as our baselines: FedAvg, Per-FedAvg, pFedMe, FedAMP and FedEM\cite{marfoq2021federated}. These baselines are respectively classical federated learning, regularization based, MAML-based meta learning, FTML methods and method combining both EM and FMTL.

\subsection{Results}
\label{ssec_expr}

\subsubsection{Hyper-Parameter Settings}
\label{sssec_hps}

The hyper-parameters used in the test of Table \ref{tbl_res_lnlm}, which are not well-tuned (parts of them from their origin paper), are the same as possible and shown as follows: the stepsize of the main problem $\alpha_{m}$ and personalized stepsize $\alpha$ for all methods are 0.01, all the $\beta$ is 1, the number of local epochs $R$ is 20 for all settings, the $\lambda$ is chosen from 15.0 to 60.0, the batchsize of Sent140 and the ones of the other datasets are 400 and 20 respectively, the aggregation strategy $\mathcal{I}$ is uniform sampling, the ratios of aggregated clients per global epoch are 40\%, 10\% and 20\% for Sent140, FEMNIST and the other datasets respectively, the numbers of total clients $N$ are 10, 198, 20 and 100 for Sent140, FEMNIST, CIFAR-10 and other datasets, the number of proximal iteration is 5 for all settings about proximal mapping and $\eta_{\alpha}$ and $\eta$ in our implementations are respectively 0.01 and 0.05.

\subsubsection{Tricks}
\label{sssec_tricks}

The related tricks are as following:
\begin{itemize}
    \item \textbf{FT}: Fine-tuning single personalized model one more step before local testing with a batch of local training data.
    \item \textbf{AM}: Aggregate momentum, the same trick used in line 14 of Algorithm \ref{alg_pFedBreD}.(To compare more fairly between methods with and without this trick; $\beta=2$ for methods with single global model and employing \text{AM})
\end{itemize}

\begin{table}[t]
    \centering
    \resizebox{.94\textwidth}{!}{
    \begin{tabular}{l|c|cc|cc|cc|c|c}
        \hline
        \textbf{Methods} & \textbf{Datasets} & \multicolumn{2}{c|}{FEMNIST} & \multicolumn{2}{c|}{FMNIST} & \multicolumn{2}{c|}{MNIST} & CIFAR-10 & Sent140\\

        & \textbf{Models} & MCLR & DNN & MCLR & DNN & MCLR & DNN & CNN & LSTM \\
                                
        \hline
        
        FedAvg & \textbf{G} & 53.96 & 57.94 & 82.90 & 81.22 & 86.96 & 88.74 & 58.27 & 71.15\\
        
        FedAvg+AM & \textbf{G} & 56.07 & 60.03 & 83.04 & 81.93 & 87.10 & 89.86 & 59.17 & 71.67\\
        
        \hline
        
        FedEM & \textbf{G} & 40.87 & 43.96 & 92.47 & 95.23 & 85.85 & 86.47 & 56.26 & 67.05\\

        FedAMP & \textbf{P} & \underline{60.04} & 66.75 & \textbf{98.72} & 98.88 & \textbf{90.89} & 92.23 & 77.33 & 69.40\\

        \hline
        
        pFedMe & \textbf{P} & 51.70 & 54.72 & 97.83 & 98.63 & 88.48 & 90.62 & 73.05 & 69.63\\
        & \textbf{G} & 50.49 & 56.20 & 82.30 & 80.16 & 86.34 & 88.22 & 43.63 & 69.82\\
        
        pFedMe+FT & \textbf{P} & 52.16 & 53.79 & 97.69 & 98.63 & 88.37 & 91.71 & 65.04 & 68.19\\
        & \textbf{G} & 50.10 & 55.89 & 81.42 & 81.11 & 88.35 & 90.75 & 42.85 & 69.48\\
        
        pFedMe+AM & \textbf{P} & 56.91 & 60.46 & 97.81 & 98.70 & 88.66 & 91.30 & 74.51 & 71.53\\
        & \textbf{G} & 54.13 & 57.70 & 83.12 & 80.54 & 86.85 & 89.53 & 47.43 & 71.43\\
        
        \hline

        Per-FedAvg  & \textbf{P} & 54.34 & 61.57 & 93.69 & 97.13 & 87.01 & 91.04 & 78.41 & 69.72\\
        & \textbf{G} & 53.30 & 56.30 & 81.86 & 72.90 & 86.72 & 90.93 & 27.22 & 70.33\\

        Per-FedAvg+FT  & \textbf{P} & 55.41 & 63.99 & 95.53 & 98.55 & 87.23 & 91.31 & 79.55 & 70.11\\
        & \textbf{G} & 53.69 & 57.12 & 81.75 & 73.24 & 86.76 & 91.02 & 28.92 & 69.77\\
        
        Per-FedAvg+AM  & \textbf{P} & 58.17 & 66.03 & 94.15 & 98.15 & 87.32 & 90.97 & 79.46 & 70.86\\
        & \textbf{G} & 56.16 & 59.34 & 81.23 & 79.04 & 86.75 & 90.91 & 36.90 & 70.64\\
           
        \hline
        
        $\mathbf{mg}$ (ours) & \textbf{P} & 56.38 & 65.29 & 98.57 & 98.88 & 89.90 & 92.07 & \underline{79.86} & \underline{72.25}\\
        & \textbf{G} & 55.38 & 60.08 & 82.58 & 79.55 & 86.91 & 89.41 & 52.66 & 72.01\\
        
        $\mathbf{mg}$ (ours)+FT & \textbf{P} & 59.82 & \underline{67.80} & \underline{98.59} & \textbf{99.06} & \underline{89.94} & \textbf{92.47} & 79.24 & 71.16\\
        & \textbf{G} & 55.10 & 60.16 & 82.90 & 82.35 & 87.03 & 90.92 & 52.03 & 71.16\\

        $\mathbf{mg}$ (ours)+AM & \textbf{P} & \textbf{60.64} & \textbf{70.34} & 98.57 & \underline{98.94} & 89.93 & \underline{92.32} & \textbf{80.63} & \textbf{73.81}\\
        & \textbf{G} & 56.23 & 59.53 & 82.57 & 79.64 & 86.92 & 90.19 & 54.09 & 73.71\\
     
        \hline
    \end{tabular}
    }
    \caption{Results of average testing accuracy (\%) per client of each settings. We mark the best and second best performance by \textbf{bold} and \underline{underline} of both global and personalized model.}
    \label{tbl_res_lnlm}
\end{table}

\subsubsection{Global and Local Test}
\label{sssec_glt}

Global test means all the test data is used in the test, local test means only the local data is used for the local test and the weight of the sum in local test is the ratio of the number of data. The results of average accuracy per client are shown in Table \ref{tbl_res_lnlm} (The maximum results of each settings are shown; More details through the optimizing processes are in Appendix \ref{appdx_sec_exp}), and each experiment is repeated 5 times. The global and personalized model, represented by $\textbf{G}$ and $\textbf{P}$, are evaluated with global and local test respectively. 

\subsubsection{Hyper-Parameter Effect}
\label{sssec_pe}

We test the hyper-parameter effect of $\eta$ and $\lambda$ in our implementation pFedBreD$_{ns,\mathbf{mg}}$ and details in Appendix \ref{appdx_sec_exp}. We observe the effects that the values of $\lambda$ or $\eta$ are too large or too small, which will degrade the test accuracy, the test accuracy of personalized model is more sensitive than the ones of global model, and to $\eta$ than to $\lambda$. Moreover, there is room for performance improvement of our implementations in Table \ref{tbl_res_lnlm} by well-tune $\eta$ and $\lambda$.

\subsection{Personalization}
\label{ssec_personalization}

\subsubsection{Deviation}

\begin{figure}[t]
    \centering
    \includegraphics[width=1.5in,height=0.8in]{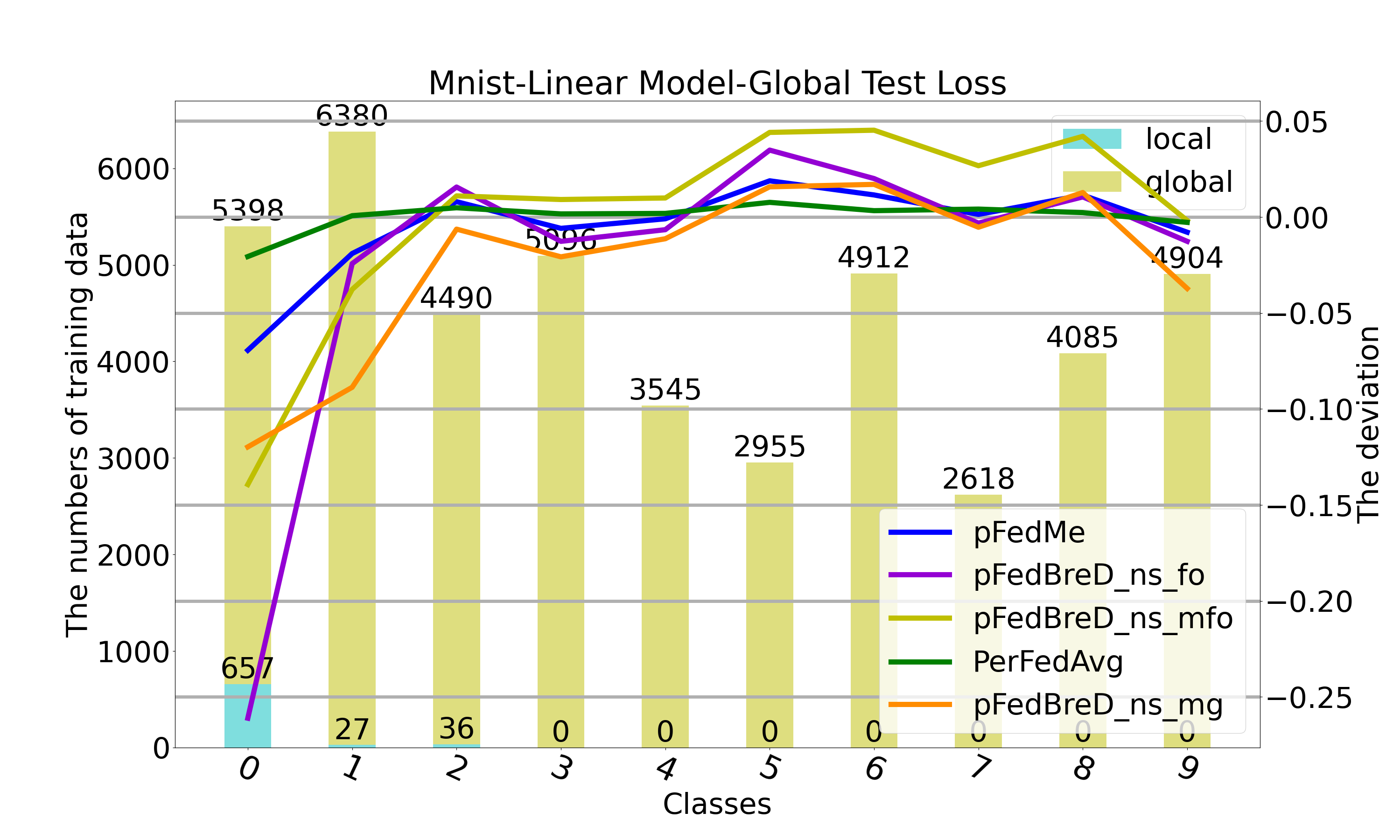}
    \includegraphics[width=1.5in,height=0.8in]{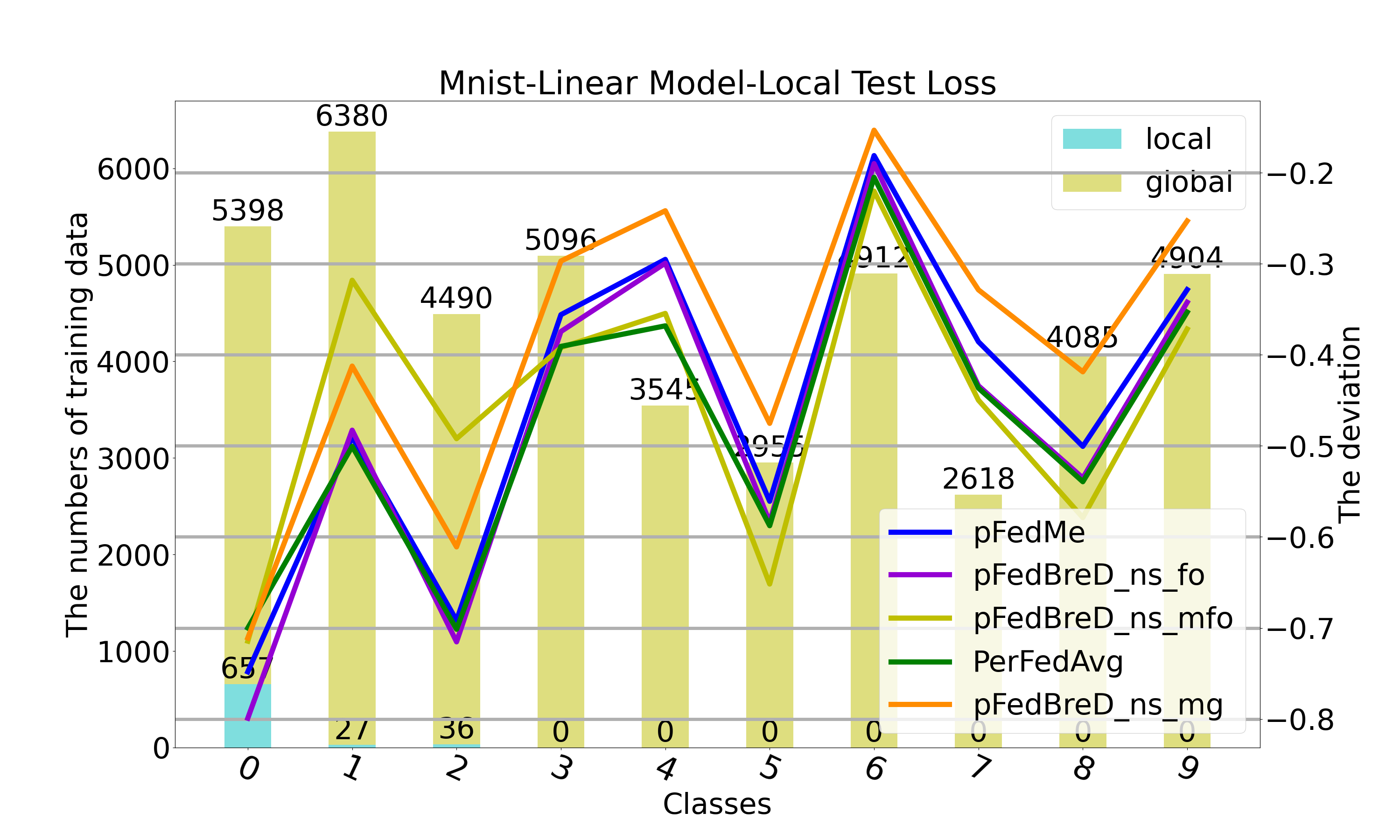}
    \includegraphics[width=1.5in,height=0.8in]{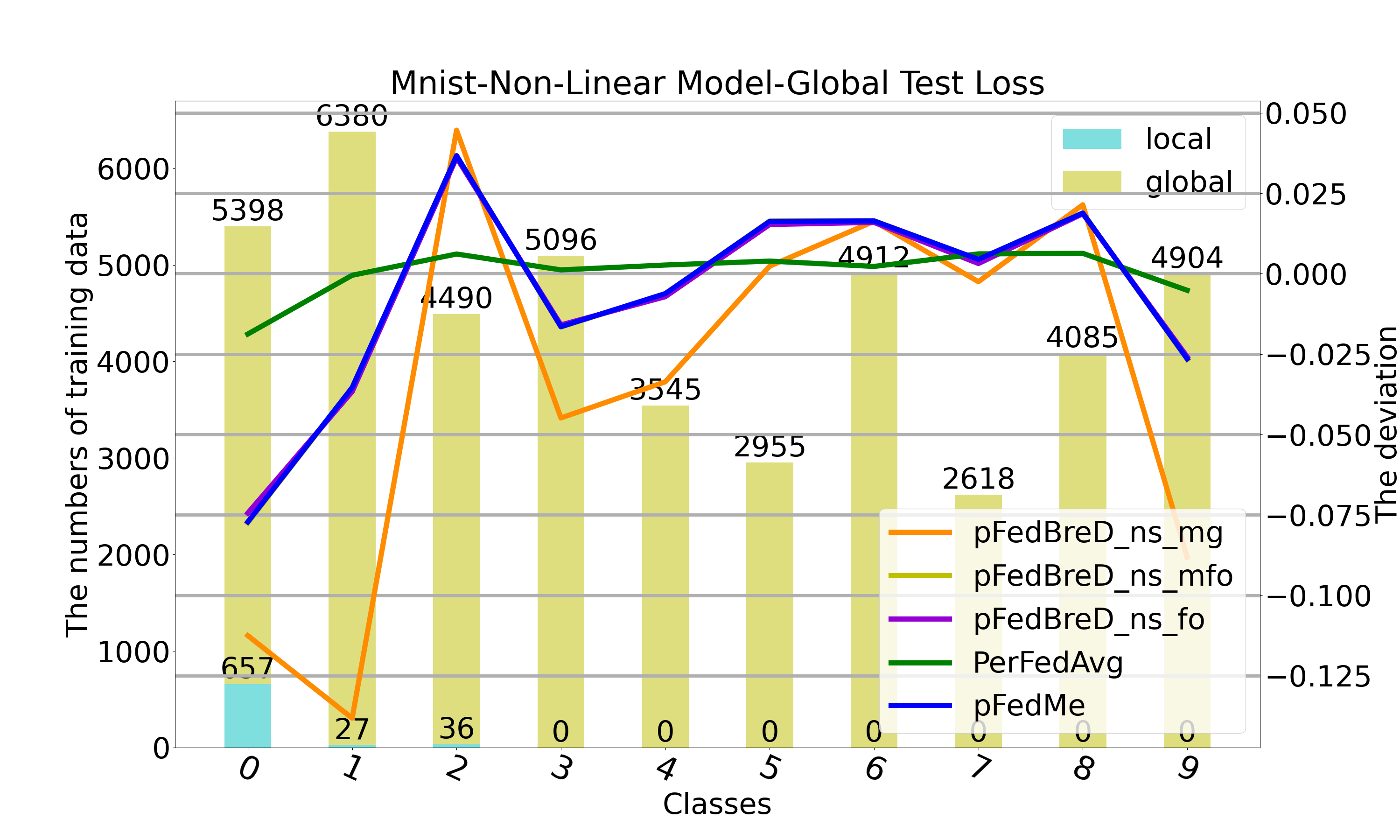}
    \includegraphics[width=1.5in,height=0.8in]{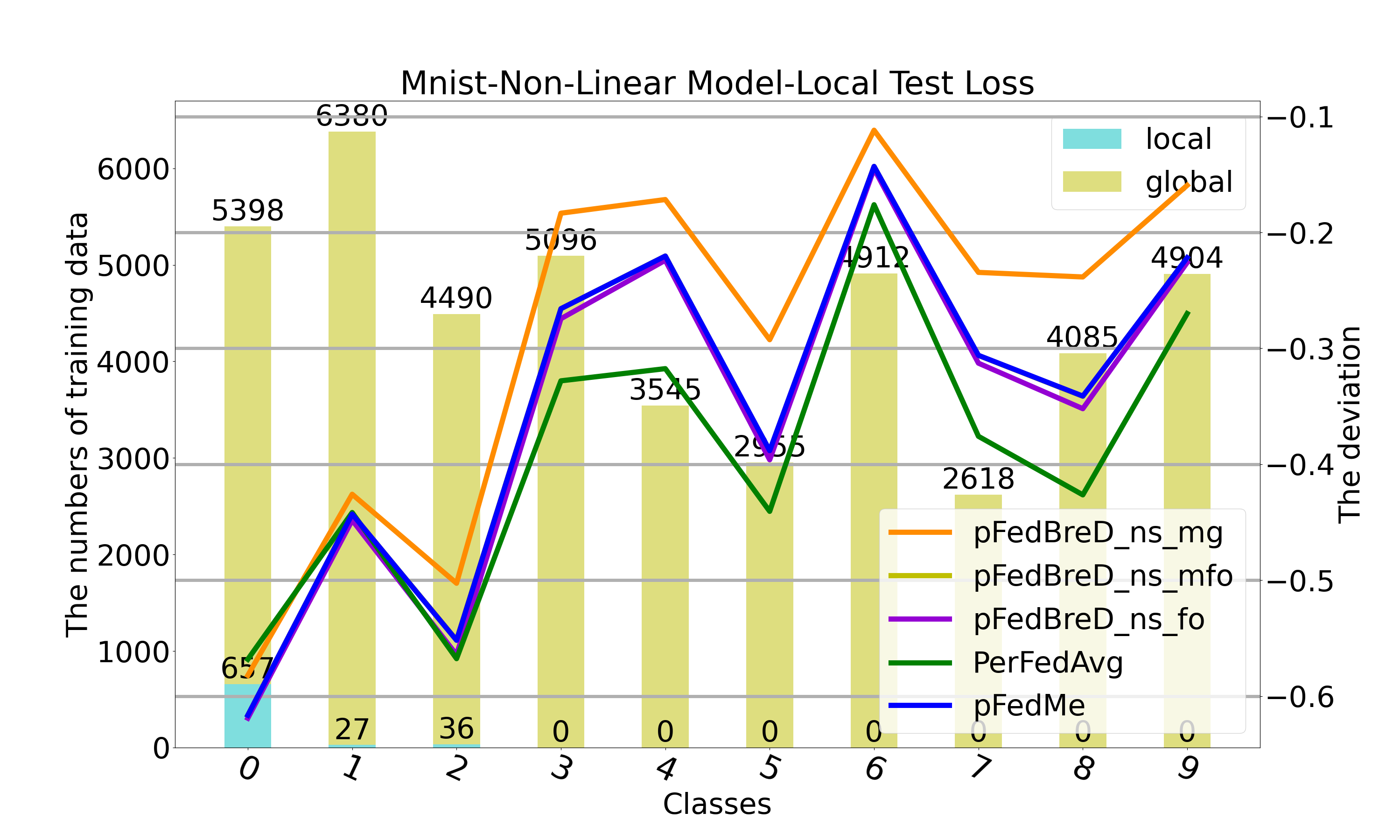}
    \begin{center}
        \footnotesize 
        MNIST-MCLR-G \qquad\qquad\qquad MNIST-MCLR-L \qquad\qquad\qquad
        MNIST-DNN-G \qquad\qquad\qquad MNIST-DNN-L
    \end{center}

    \caption{The loss deviation of our experiments in Section \ref{ssec_expr} on the first client. The histogram represent the number of both global data and local data on the first client with each class of label. The folding lines represent each personalized methods' $\Delta \mathcal{G}_{1,c}$ and $\Delta \mathcal{G}_{1,c}$, the global (G) and local (L) test loss deviation over data with each class of label on the first client, where $c$ is the integer from 0 to 9.}
    \label{fig_deviation}
\end{figure}

The notations are shown as follows:
\begin{itemize}
    \item $\mathcal{L}_{i,c}$: The averaged local test loss of the $i$-th personalized model over its own local test data with label $c$. The value equals zero on the clients without data with label $c$.
    \item $\bar{\mathcal{L}}_{c}$: The mean of the averaged local test loss over all personalized models. Each $\mathcal{L}_{i,c}$ is weighted by the the ratio of the number of own test data with label $c$.
    \item $\mathcal{G}_{i,c}$: The averaged global test loss of the $i$-th personalized model over the global test data with label $c$.
    \item $\bar{\mathcal{G}}_{c}$: The mean of the averaged global test loss over all personalized models
    \item The deviations of the averaged global and local test loss of the $i$-th personalized model on class $c$: $\Delta \mathcal{G}_{i,c} = \mathcal{G}_{i,c} - \bar{\mathcal{G}_{c}}$ and $\Delta \mathcal{L}_{i,c} = \mathcal{L}_{i,c} - \bar{\mathcal{L}_{c}}$..
\end{itemize}

\subsubsection{Analysis}

The deviations of one specific client in our experiments of MNIST are shown in Figure \ref{fig_deviation}(more in Appendix \ref{appdx_sec_exp}). The global test verifies the personalization ability of our implementations. The personalized models of our implementations focus on the class of locally available labels. The deviation on $c^{*}$ the class of locally unavailable labels $\Delta \mathcal{L}_{i,c^{*}}$ equals $ -\bar{\mathcal{L}}_{c^{*}}$, due to $\mathcal{L}_{i,c^{*}}=0$. Thus, from Figure \ref{fig_deviation}, we observe that $\bar{\mathcal{L}}_{c^{*}}$ of most of our implementations are lower than the ones of the others. Therefore, the values of $\mathcal{L}_{i,c_{i}}$ of our implementations should be lower than the others, $c_{i}$ the available class on $i$-th client, since the observed $\Delta \mathcal{L}_{i,c_{i}}$ are similar. Thus, our methods bring lower $\mathcal{L}_{i,c_{i}}$.

\subsection{Ablation}
\label{ssec_ablation}

\begin{table}[t]
    \centering
    \resizebox{.94\textwidth}{!}{
    \begin{tabular}{l|c|cc|cc|cc|c|c}
        \hline
        
         \textbf{Methods} & \textbf{Datasets} & \multicolumn{2}{c|}{FEMNIST} & \multicolumn{2}{c|}{FMNIST} & \multicolumn{2}{c|}{MNIST} & CIFAR-10 & Sent140\\

          & \textbf{Models} & MCLR & DNN & MCLR & DNN & MCLR & DNN & CNN & LSTM \\

         \hline
         
         $\mathbf{fo}$ & \textbf{P} & 49.39 & 49.30 & 97.78 & 98.34 & 88.30 & 90.42 & 65.26 & 60.87\\
         & \textbf{G} & 50.16 & 19.74 & 81.64 & 74.49 & 86.11 & 88.06 & 22.53& 60.43\\

        \hline
         
         $\mathbf{mfo}$ & \textbf{P} & 51.28 & 54.48 & 97.85 & 98.63 & 88.46 & 90.61 & 73.05 & 69.59\\
          & \textbf{G} & 50.16 & 56.07 & 82.17 & 80.04 & 86.30 & 88.16 & 43.55 & 69.69\\
        \hline
    \end{tabular}
    }
    \caption{Average testing accuracy (\%) in Ablation.}
    \label{tbl_res_ablation}
\end{table}

The relationship among the three prior mean selection strategies mentioned in Equation (\ref{equ_def_spmmdl}) is that the strategy $\mathbf{mg}$ is consist of $\mathbf{fo}$ and $\mathbf{mfo}$. Thus, the three implementations of pFedBreD naturally form a set of ablation experiments. 

Moreover, the performance of pFedBreD$_{ns}$ methods, as prior assumption of the spherical Gaussian is taken, is similar to the one of pFedMe if $\eta$ or $\eta_{\alpha}$ is small. Essentially, pFedMe can be regarded in our framework as the one which uses $\mu_{i}=I$ and takes the spherical Gaussian. as prior

The results of FEMNIST in Table \ref{tbl_res_ablation} reveal the instability of our implementation $\mathbf{fo}$ on small aggregation ratio, 10\%, and the relative robustness of $\mathbf{mg}$, which shows remarkable performance in most of the settings. Thus, the addition of the strategy $\mathbf{mfo}$ on $\mathbf{fo}$, reduces the instability problem of $\mathbf{fo}$ observed from results shown in Table \ref{tbl_res_ablation}.

\section{Discussion}
\label{sec_disc}

\subsection{Limitation}
\label{ssec_limitation}

The instability still exists in the implementation $\mathbf{mg}$ in the settings of small aggregation ratio.(Appendix \ref{appdx_ssec_instability}) The superficial reason is that $\eta_{\alpha}$ and $\eta$ (similar to each other) are used simultaneously, making the magnitude in $\mathbf{mg}$ twice as the ones in the other two implementations. Empirically, twice the $\eta_{\alpha}$ in $\mathbf{fo}$ doesn’t improve the performance, but one more $\mathbf{fo}$ step used in $\mathbf{mfo}$ brings significant improvement as $\mathbf{mg}$. We believe that more theoretical findings are needed to support the analysis, and one possible theoretical explanation we believe is that the gradient of the local loss function derived based on the current global model is rather unstable, and this instability may arise from the non-i.i.d. in FL settings and extremely unbalanced data.

As shown in Table \ref{tbl_res_lnlm}, although the performance of $\mathbf{mg}$ can be improved to some extent by tricks on non-convex problems, the performance is slightly worse than FedAMP, which can also be seen as a special prior selection strategy in our framework, on convex problems.\cite{huang2021personalized}

\subsection{Prior Selection}
We use the exponential family because it is relatively large and has many computationally convenient properties. Other families of distributions are also worth considering, such as long-tailed ones, and these may induce more optimization frameworks to accommodate a wider range of situations (e.g. different assumptions or realities of data distributions).

\subsection{Complexity}

Since the general process of our implementations, FedAMP and pFedMe are the same as shown in pFedBreD framework, these methods share the same complexity of memory/calculation, $O(N)/O(NTRK)$.
The complexities of both FedAvg and Per-FedAvg are $O(N)/O(NTR)$ since the original methods of them do not need a approximate proximal mapping solution, and therefore are free on $K$, the number of iterations to calculate the solution.
The complexity of FedEM is $O(NM)/O(NTRM)$, where $M$ is the components of the distributions we assume, due to the calculation of $M$ components in each global epoch.

\subsection{Bayesian Neural Network}

Our baselines do not contain BNN-based methods such as pFedBayes, mainly because it is difficult to conduct comparative experiments by fixed some components (e.g., inferential models, tricks and optimization methods), and our method is, strictly, not an variational inference method, but a point estimation and regularization-based optimization method based on expectation propagation \cite{Minka2001ExpectationPF,Seeger2009ExpectationPF}. If we have had them in Table \ref{tbl_res_lnlm}, it is hard to discuss which part contributes the most, and the inferential model is far different from the others including ours.

\section{Conclusion}
\label{sec_cncls}

This work has several contributions as follows: We propose pFedBreD a unified framework to solve a class of MAP problem for personalization in the federated learning setting, which is modeled with Bayesian learning under prior assumption of distribution in SX-family; 
We propose the personalized prior methods with 3 optional selection strategies;
We empirically verify the performance of 3 first order implementations of pFedBreD under prior assumption of the spherical Gaussian and analyze their personalization.

\bibliography{pFedBreD.bib}
\bibliographystyle{unsrt}

\appendix
\setcounter{equation}{0}
\setcounter{figure}{0}
\setcounter{table}{0}

\newpage
\newpage

\section{Glossary and Some Basic Knowledge}

\subsection{Glossary}

The main notations in this paper are shown in Table \ref{appdx_tbl_glsry}.

\begin{table}[ht]
    \centering
    \resizebox{.94\textwidth}{!}{
    \begin{tabular}{c|c}
        \hline
        Notation & Implication \\
        \hline
        $\cdot_{i}$ & $\cdot$ on $i$-th client \\
        $f_{i}$ & local loss function \\
        $F_{i}$ & local objective function \\
        $F$ & global objective function \\
        $\mathbf{E}_{\cdot}$ & expectation on $\cdot$ \\
        $\mathbf{H}$ & entropy \\
        $\mathbf{P}$ & probability measure \\
        $\mathbf{P}_{ef}$ & probability in exponential family\\
        $\mathbf{P}_{sef}$ & probability in scaled exponential family \\
        $\Omega_{\cdot}$ & complete set of $\cdot$ \\
        $\mathcal{F}$ & generic $\sigma$-algebra\\
        $\sigma(\cdot)$ & $\sigma$-algebra derived from $\cdot$ \\
        $\{\Omega,\mathcal{F}\}$ & measurable space \\
        $\{\Omega,\sigma(\Omega),\mathbf{P}\}$ & probability measurable space \\
        $\hat{\mathbf{P}}$ & estimated probability \\
        $x_{i}$,$y_{i}, d_{i}$ & input data, label data, the pairs of them \\
        $w$ & global model parameter \\
        $\Theta_{i}$ & local information \\
        $\theta_{i}$ & personalized parameters\\
        $w_{init},\theta_{init}$ & function to initialize parameters \\
        $\mu_{i}$ & the function to generate mean parameter \\
        $s_{i}$ & the function to generate natural parameter \\
        $x$,$y$,$\hat{w}$,$\theta$,$\mu$,$s$ & generic point notations \\
        T,N,R & number of total global epochs, clients, local epochs \\
        t,r & global epochs, local epochs \\
        $\beta$,$\eta$,$\eta_{\alpha}$,$\lambda$,$\hat{\lambda}$ & scalar notations \\
        $g,h,h_{\lambda}$ & generic function notations \\
        $\mathcal{D}_{g}$ & Bregman divergence derived from g\\
        $\mathcal{D}\mathbf{prox}$ & Bregman divergence proximal mapping \\
        $\mathcal{D}\mathbf{env}$ & Bregman Moreau envelope \\
        $\nabla, \mathbf{D}, \nabla^{2}$ & gradient, Jocobian and Hessian operator \\
        $\Delta$ & deviation from mean \\
        $\cdot^{*}$ & the Fenchel conjugate of $\cdot$ \\
        $\mathcal{L}$ & averaged local test loss \\
        $\mathcal{G}$ & averaged global test loss \\
        $\bar{\cdot}$ & mean of $\cdot$ over clients \\   
    \end{tabular}}
    \caption{The glossary of notations mentioned in this paper}
    \label{appdx_tbl_glsry}
\end{table}

\subsection{Bregman Divergence}

Bregman divergence is a general distance satisfying that its first-order moment estimation is the point that minimizes the expectation of the distance to all points for all measurable functions on $\mathbf{R}^{d}$. In other words, the given distance $\mathcal{D}$ satisfies Condition (\ref{bd_prpt}):
\begin{equation}
\label{bd_prpt}
 \forall X \in \{\mathbf{R}^{d}, \mathcal{F}, \mathbf{P}\}, \mathbf{E}[X]={\arg\min}_{y}\mathbf{E}[\mathcal{D}(X,y)   ]
\end{equation}

Equation (\ref{def_bd}) is the definition of Bregman divergence:
\begin{equation}
\label{def_bd}
\begin{aligned}
    \mathcal{D}_{g}(x,y)&:=g(x)-g(y)-\langle\nabla g(y),x-y\rangle \\ 
    &=\int_{y}^{x} \nabla g(t)-\nabla g(y) dt
\end{aligned}
\end{equation}
where $g$ is a convex function. For convenience, in this paper, $g$ is assumed to be strictly convex, proper and differentiable, so that the equation above Equation (\ref{def_bd}) are well-defined. In the perspective of Taylor expansion, Bregman divergence is the first order residual of $g$ expanded at point $y$ valued at point $x$, which is the natural connection between Bregman divergence and Legendre transformation. The Bregman divergence does not satisfy the distance axiom, but it provides some of the properties we need, such as non-negative distance. Hence, the selected function $g$ should be convex. Furthermore, if one wants the distance to have a good property that $x=y\leftrightarrow\mathcal{D}_{g}(x,y)=0$, one needs $g$ to be strictly convex.

\subsection{Sampling Method in Bayesian Learning}

The sampling methods used to calculate the solution of Bayesian Model mentioned in this paper can be importance sampling, MCMC or others. In this work, we use the approximation Bayesian methods.(See more details in \cite{andrieu2003introduction})

\section{Details of Equations}
\label{appdx_sec_doe}
\subsection{Hidden Information}

From the definition of KL divergence, we have 
\begin{equation}
\begin{aligned}
&\arg\min_{w}\mathbf{E}_{i}\mathbf{E}_{d_{i}}\mathbf{KL}(\mathbf{P}(y_{i}|x_{i})||\hat{\mathbf{P}}(y_{i}|x_{i},w)) \\
=&\arg\min_{w}\mathbf{E}\log \mathbf{P}(y_{i}|x_{i}) - \log \hat{\mathbf{P}}(y_{i}|x_{i},w)) \\
=&\arg\min_{w}\mathbf{E} - \log \hat{\mathbf{P}}(y_{i}|x_{i},w)) \\
=&\arg\max_{w}\mathbf{E} \log \hat{\mathbf{P}}(y_{i}|x_{i},w)) \\
\end{aligned}
\end{equation}
This is used in Equation (\ref{equ_gfl}) in the main paper.

\subsection{Bregman Divergence and X-Family}

If proper and strictly convex function $g$ is differentiable, with $g^{*}$ the Fenchel conjugate function of $g$, $\mathcal{D}_{g}(x,y)$ the Bregman divergence, $\mu$ dual point of $s$, we have:
\begin{equation}
    \begin{aligned}
        \mathcal{D}_{g^{*}}(\mathcal{V},\mu)=g^{*}(\mathcal{V})+g(s)- \langle \mathcal{V}, s \rangle=\mathcal{D}_{g}[s, \nabla g^{*}(\mathcal{V})]
    \end{aligned}
\end{equation}

From the definition of Bregman divergence , $\nabla g(s) = \mu$ and definition of $g^{*}$ Fenchel conjugate on convex function $g$ ,we have:
\begin{equation}
    \begin{aligned}
        \mathcal{D}_{g^{*}} (\mathcal{V}, \mu)
        &= g^{*}(\mathcal{V}) - g^{*}(\mu) - \langle \nabla g^{*}(\mu), \mathcal{V} - \mu \rangle \\
        &= g^{*}(\mathcal{V}) - g^{*}(\mu) - \langle s, \mathcal{V} - \mu \rangle \\
        &= g^{*}(\mathcal{V}) - \langle s, \mathcal{V} \rangle - g^{*}(\mu) + \langle \mu, s \rangle \\
        &= g^{*}(\mathcal{V}) - \langle s, \mathcal{V} \rangle + g(s) \\
    \end{aligned}
\end{equation}
Similarly, we have $\mathcal{D}_{g}[s, \nabla g^{*}(\mathcal{V})] = g
^{*}(\mathcal{V}) - \langle s, \mathcal{V} \rangle + g(s)$. This property is used in Equation (\ref{equ_def_ef}) and (\ref{equ_def_sef}) in the main paper.

\begin{table}[ht]
    \centering
    \resizebox{.94\textwidth}{!}{
    \begin{tabular}{ccccc}
        \toprule
         Name & Gaussian & Bernoulli & Possion & Exponential \\
         \midrule
          Domain& $\mathbf{R}^{d}$ & $\{0,1\}$ & $\mathbf{N}$ & $\mathbf{R}_{++}$\\
          $g(y)$ & $\frac{1}{2}||y||_{\Sigma^{-1}}^{2}$ & $\ln(1+e^{y})$ & $e^{y}$ & $-\ln(-y)$\\
          $\nabla g(y)$ & $y$ & $\frac{\exp\{y\}}{1 + \exp\{y\}}$ & $e^{y}$ & $-y^{-1}$\\
          $g^{*}(x)$ & $\frac{1}{2}||x||_{\Sigma^{-1}}^{2}$ & $\xi(x)+\xi(1-x)$ & $x\ln(x)-x$ & $-\ln(x)-1$ \\
          $\nabla g^{*}(x)$ & $x$ & $\ln(\frac{x}{1-x})$ & $\ln(x)$ & $-x^{-1}$ \\
          $\mathcal{D}_{g^{*}}(x,y)$ & $\frac{1}{2}||x-y||_{\Sigma^{-1}}^{2}$& $\ln(1+e^{(1-2x)y})$ & $e^{y}+\xi(x)-x(y+1)$ & $\frac{x}{y}-\ln\frac{x}{y}-1$\\
         \bottomrule
    \end{tabular}
    }
    \caption{Bregman divergence and exponential family. (note $\xi=\langle\cdot ,\ln\cdot\rangle$)}
    \label{tbl_bd_ef}
\end{table}

Table \ref{tbl_bd_ef} shows parts of the relationship between specific $g$ and related member in exponential family. See \cite{banerjee2005clustering} for more about the relationships between $g$ that derives Bregman divergence $\mathcal{D}_{g}$ and related derived divergence (e.g., $\cdot\Sigma^{-1}\cdot$ \& Mahalanobis distance, $\sum_{\cdot} \cdot \log \cdot$ \& KL divergence / generalized I-divergence and etc.).

\subsection{Expectation Maximization}
\label{appdx_ssec_em}
The details of Equation (\ref{equ_gfl_eml}) in the main paper is shown in Equation (\ref{appdx_equ_gfl_eml}).
\begin{equation}
\label{appdx_equ_gfl_eml}
\begin{aligned}
& \sum_{i}\log\mathbf{P}(y_{i}|x_{i},w) = \sum_{i}\log\int\mathbf{P}(y_{i},\Theta_{i}|x_{i},w)d\Theta_{i} \\
=& \sum_{i}\log\int\mathbf{Q}(\Theta_{i})\frac{\mathbf{P}(y_{i},\Theta_{i}|x_{i},w)}{\mathbf{Q}(\Theta_{i})}d\Theta_{i} \\
\ge& \sum_{i}\int\log\mathbf{Q}(\Theta_{i})\frac{\mathbf{P}(y_{i},\Theta_{i}|x_{i},w)}{\mathbf{Q}(\Theta_{i})}d\Theta_{i} \\
=& \sum_{i}\mathbf{E}_{\mathbf{Q}(\Theta_{i})}\log\frac{\mathbf{P}(y_{i},\Theta_{i}|x_{i},w)}{\mathbf{Q}(\Theta_{i})} \\
=& \sum_{i}\mathbf{E}_{\mathbf{Q}(\Theta_{i})}\log\mathbf{P}(y_{i},\Theta_{i}|x_{i},w)-\log\mathbf{Q}(\Theta_{i}) \\
\ge& \sum_{i}\mathbf{E}_{\mathbf{Q}(\Theta_{i})}\log\mathbf{P}(y_{i},\Theta_{i}|x_{i},w)\\
=& \sum_{i}\mathbf{E}_{\mathbf{Q}(\Theta_{i})}[\log\hat{\mathbf{P}}(y_{i}|x_{i},\Theta_{i},w)+\log\mathbf{P}(\Theta_{i}|x_{i},w)
] \\
=& \sum_{i}\mathbf{E}_{\mathbf{Q}(\Theta_{i})}[\log\hat{\mathbf{P}}(y_{i}|x_{i},\Theta_{i},w)+\log\int_{y_{i}}\mathbf{P}(\Theta_{i}|d_{i},w)\mathbf{P}(y_{i}|x_{i},w)
] \\
\ge& \sum_{i}\mathbf{E}_{\mathbf{Q}(\Theta_{i})}[\log\hat{\mathbf{P}}(y_{i}|x_{i},\Theta_{i},w)+\mathbf{E}_{y_{i}|x_{i},w}\log\mathbf{P}(\Theta_{i}|d_{i},w)
]
\end{aligned}
\end{equation}
In Equation \ref{appdx_equ_gfl_eml}, we use the concavity of logarithmic function for the first inequality and entropy $\mathbf{H}(\mathbf{Q}(\Theta_{i}))=\mathbf{E}_{\mathbf{Q}(\Theta_{i})}-\log\mathbf{Q}(\Theta_{i}) \ge 0$ the for the second. (probability $\mathbf{Q}(\Theta_{i})\in[0,1]$; The first equal sign holds, when $\mathbf{Q}(\Theta_{i})=\mathbf{P}(\Theta_{i}|d_{i},w)$.) The last inequality is derived from the concavity of the logarithmic function.

\begin{equation}
\label{appdx_max_prpty}
\begin{aligned}
    \max_{x,y} f(x,y) &\ge \max_{x}\max_{y} f(x,y)\\
    \sum_{i}a_{i}\max f(x,y_{i}) &= \max\sum_{i}a_{i} f(x,y_{i})
\end{aligned}
\end{equation}

In Equation (\ref{equ_gfl_sembme}), we use the two properties of $\max$ shown in Equation (\ref{appdx_max_prpty}). Moreover, these properties are also used to build the upper bound of Equation (\ref{equ_gfl_md}) as Equation (\ref{equ_gfl_mdlb}).

\section{More About Experiments}
\label{appdx_sec_exp}

The access of all data and code is available

\footnote{not shown in this submission}
.

\subsection{Implementations of Per-FedAvg}

We implement Per-FedAvg with the first order method \cite{fallah2020personalized} and fine-tune the personalized model twice, with each learning step of the global and personalized step sizes.

\subsection{Non-I.I.D Distribution}

\begin{figure*}[ht]%
    \centering
    \includegraphics[width=0.16\textwidth,height=0.8in]{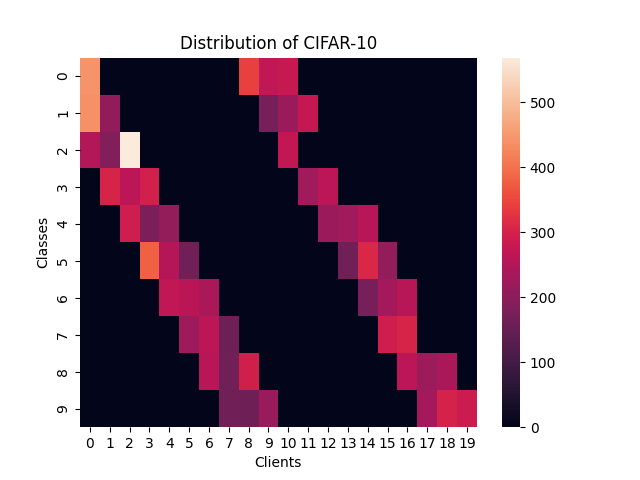}
    \includegraphics[width=0.16\textwidth,height=0.8in]{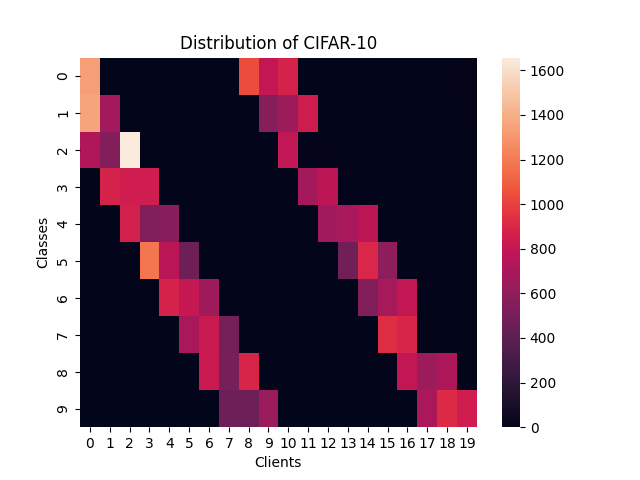}  
    \includegraphics[width=0.16\textwidth,height=0.8in]{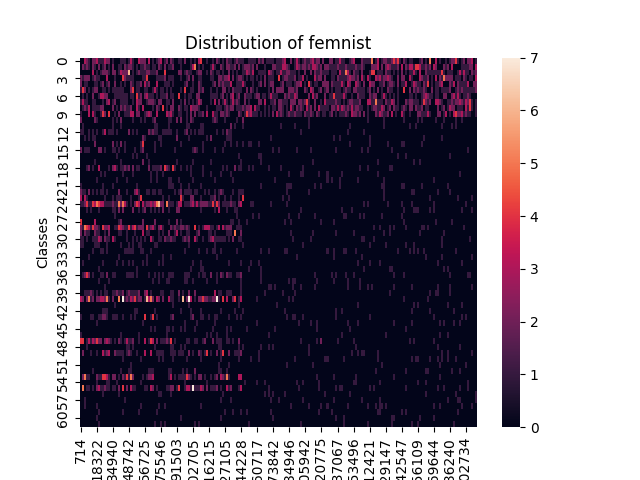}    \includegraphics[width=0.16\textwidth,height=0.8in]{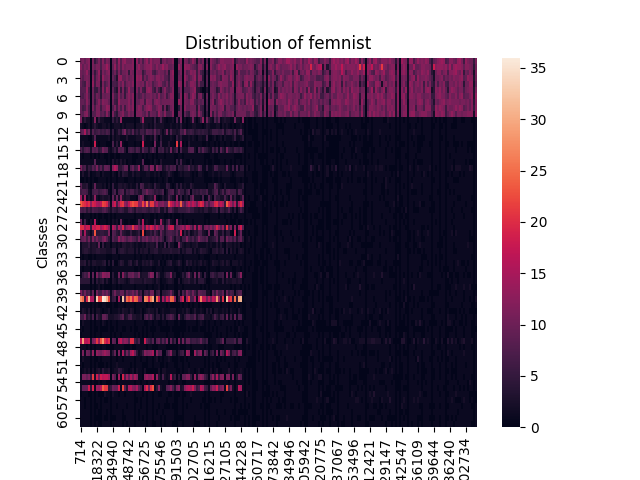}
    \includegraphics[width=0.16\textwidth,height=0.8in]{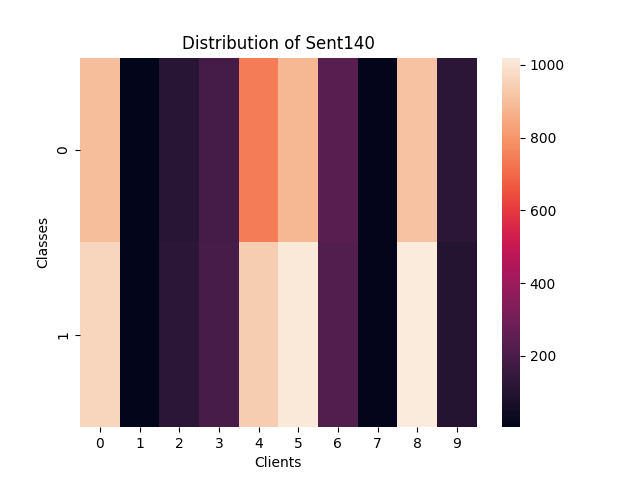}
    \includegraphics[width=0.16\textwidth,height=0.8in]{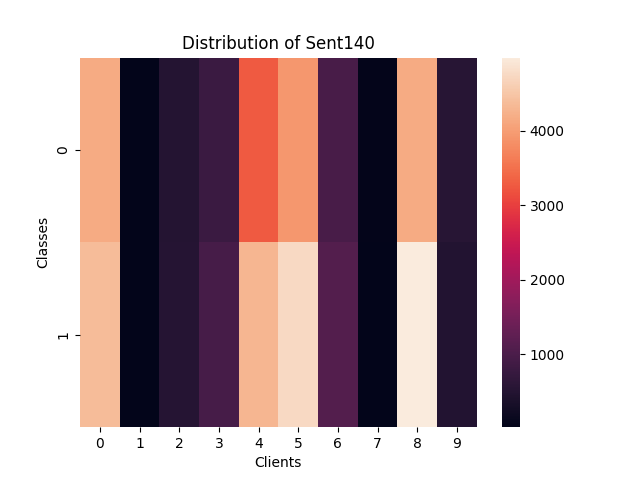}  
    \begin{center}
        \footnotesize 
        CIFAR-10-Test \qquad CIFAR-10-Train \qquad
        FEMNIST-Test \qquad FEMNIST-Train \qquad
        Sent140-10-Test \qquad Sent140-10-Train
    \end{center}
    \caption{The visualization of the non-i.i.d. data distributions of CIFAR-10, FEMNIST and Sent140.}
    \label{appdx_fig_niid}
\end{figure*}

Figure \ref{appdx_fig_niid} shows the non-i.i.d. distribution of CIFAR-10, FEMNIST and Sent140. Sent140 is a bi-level classification so each client has two class of label data and we directly use the LEAF benchmark \cite{caldas2018leaf} and Dirichlet distribution of $\alpha=0.5$ to separate users into 10 groups (See the code for more details).

\subsection{More About Hyper-Parameter Effect}

\begin{figure*}[ht]
    \centering
    \includegraphics[width=0.24\textwidth,height=0.8in]
    {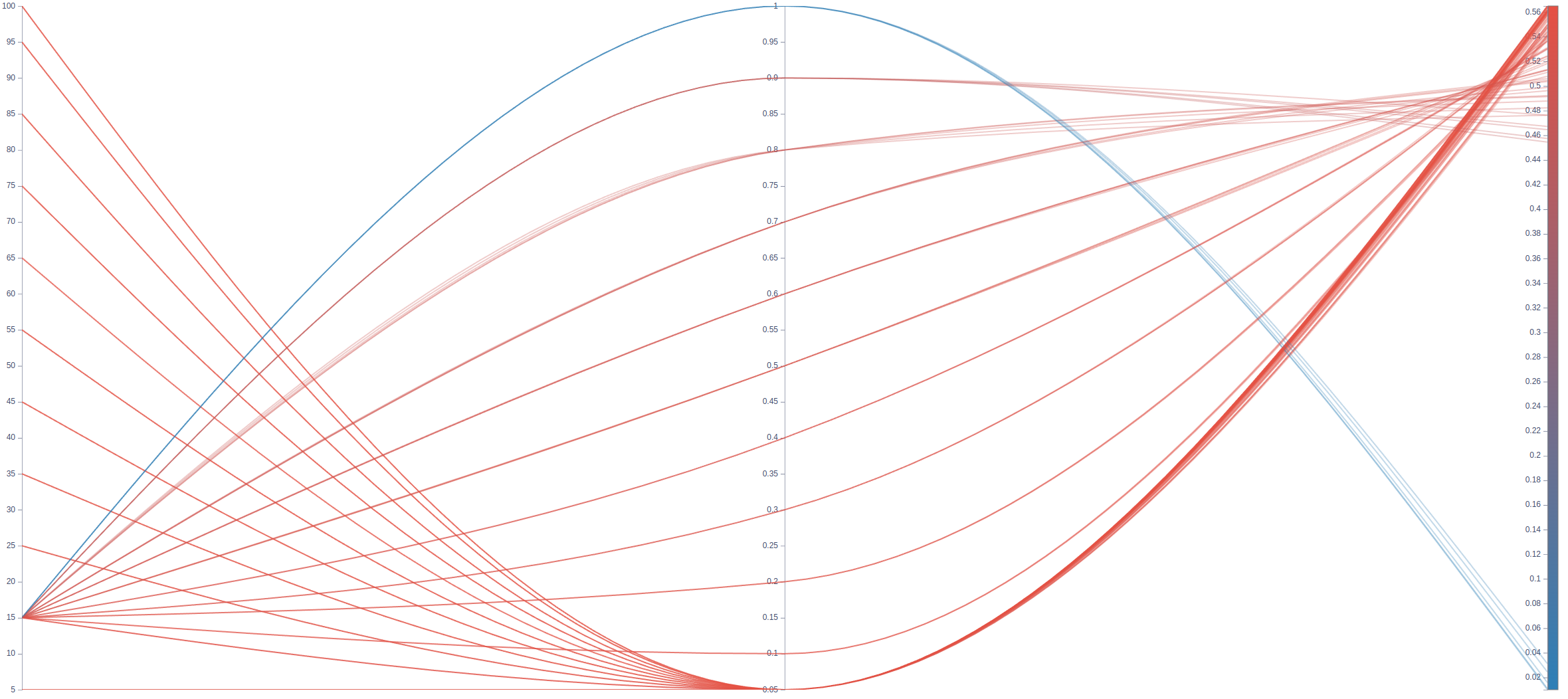}
    \includegraphics[width=0.24\textwidth,height=0.8in]
    {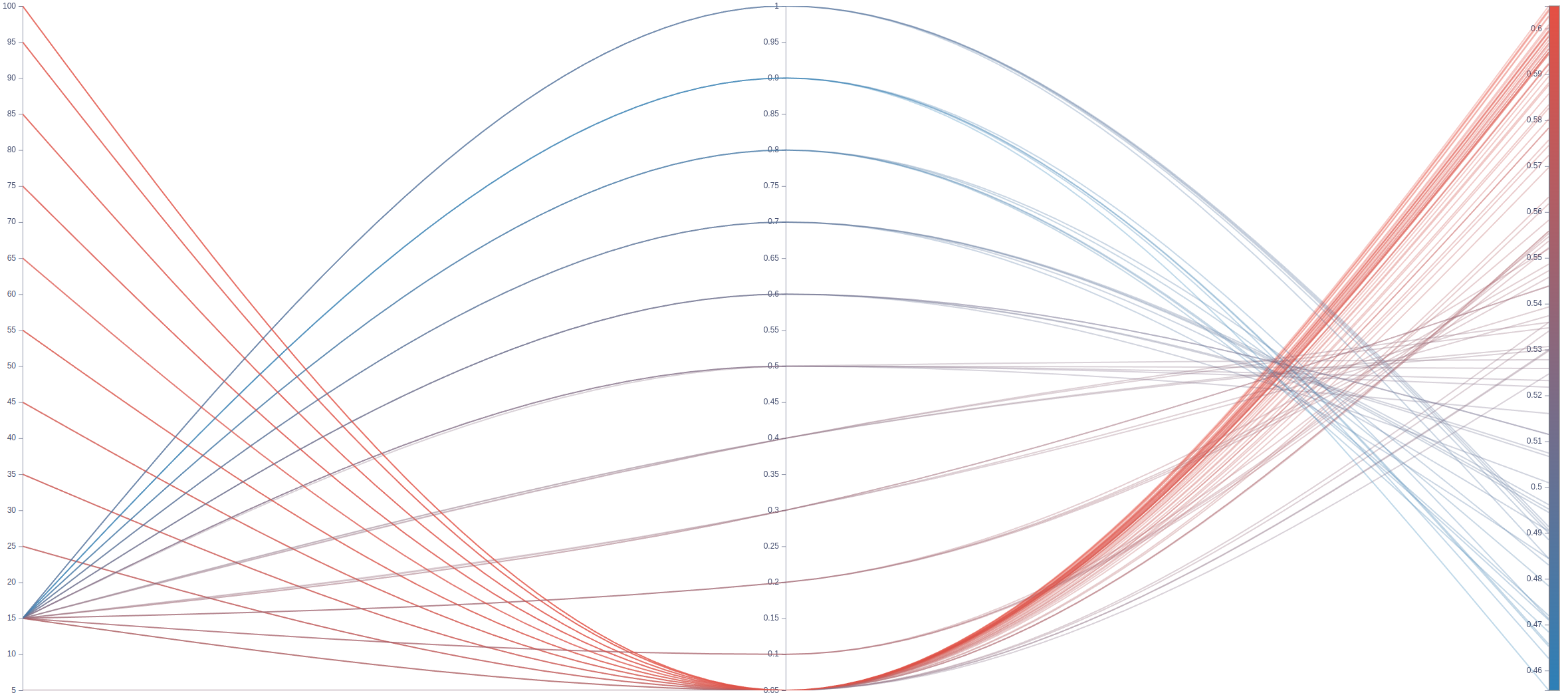}
    \includegraphics[width=0.24\textwidth,height=0.8in]{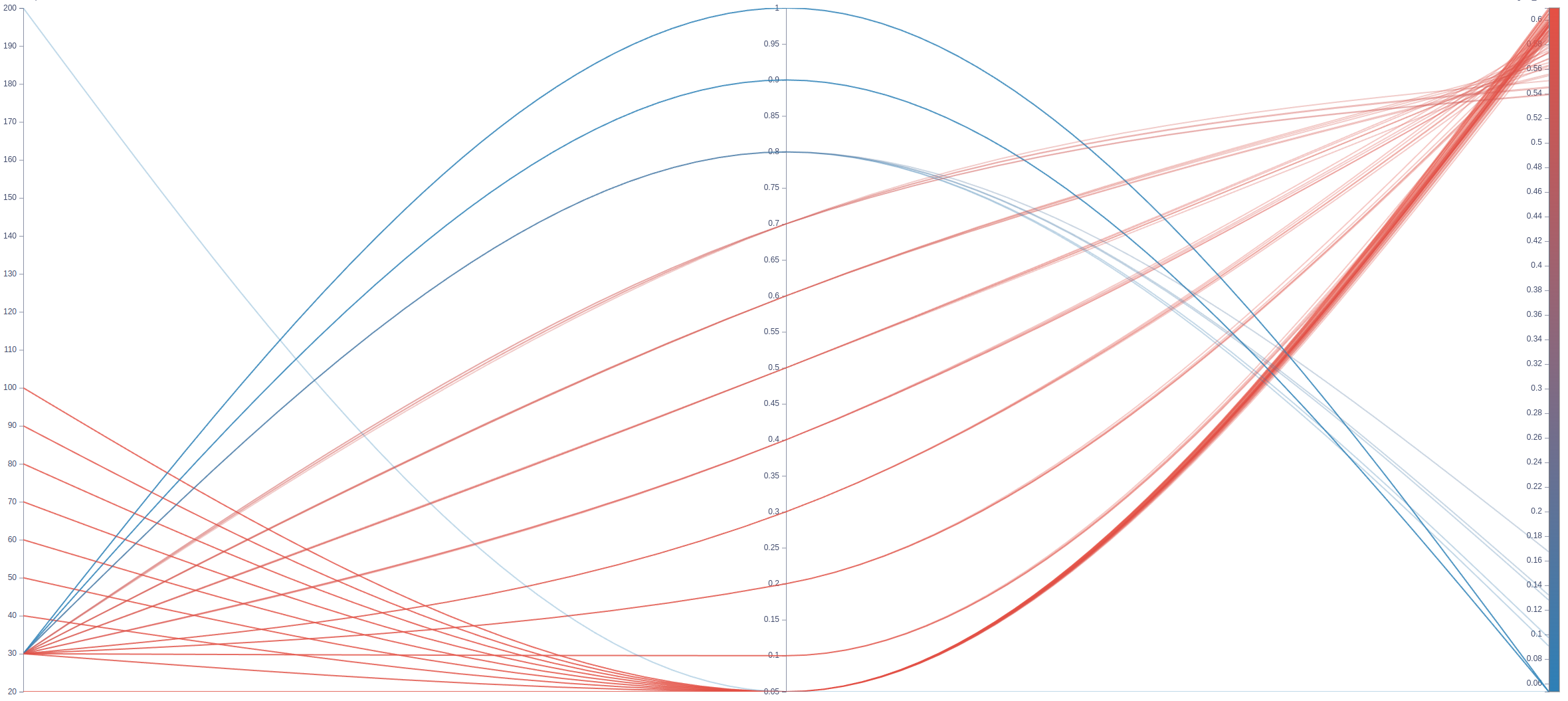}
    \includegraphics[width=0.24\textwidth,height=0.8in]{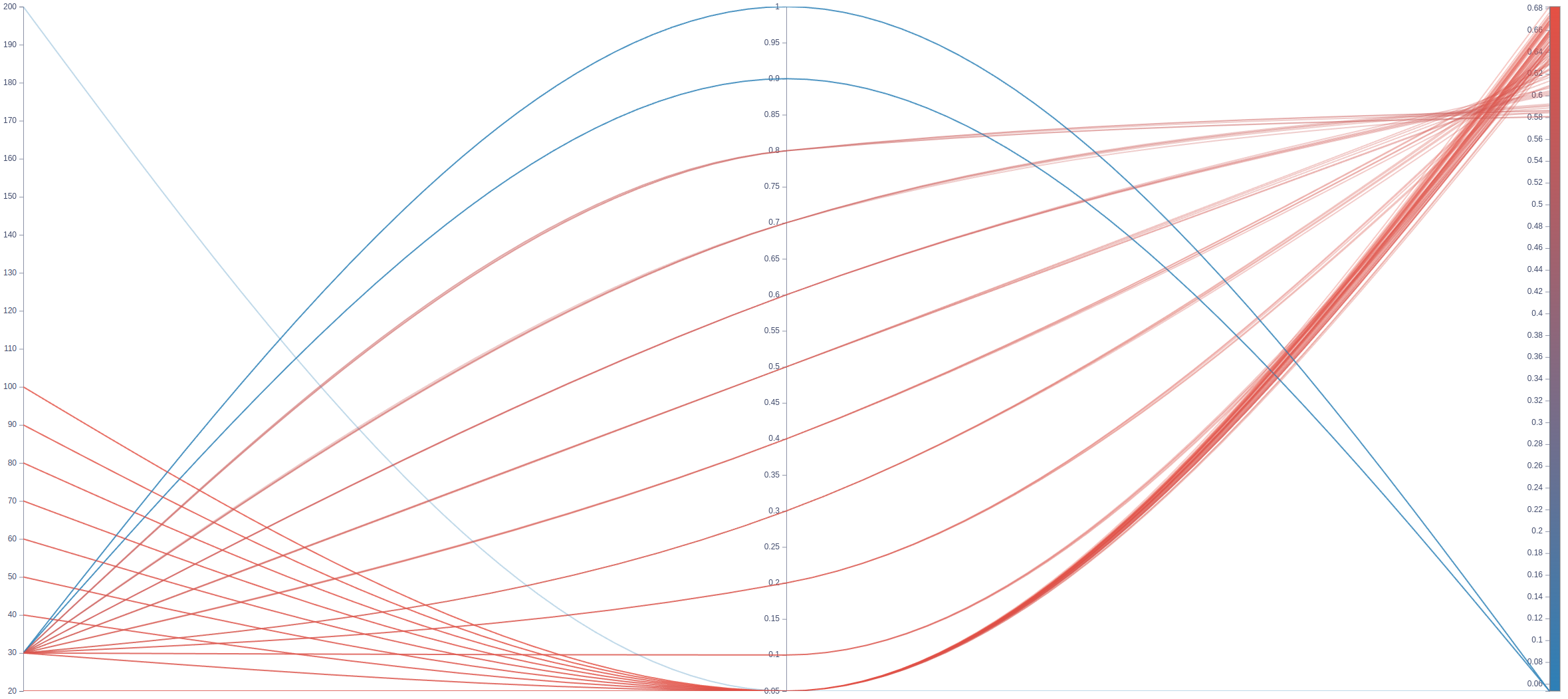}
    \begin{center}
        \footnotesize 
        FEMNIST-MCLR-G \qquad\qquad\qquad FEMNIST-MCLR-P
        \qquad\qquad\qquad\qquad
        FEMNIST-DNN-G \qquad\qquad\qquad FEMNIST-DNN-P
    \end{center}
    
    \caption{Hyper-parameter effect: The left, middle and right bars in each figure respectively represent $\lambda$, $\eta$ and test accuracy, ranges of which are respectively [0,100], [0,1] and [0,1] increasing from bottom to top (color from blue to red refers to the accuracy from 0 to 1).}
    \label{appdx_fig_hpe_femnist}
\end{figure*}

\begin{figure*}[ht]%
    \centering
    \includegraphics[width=0.24\textwidth,height=0.8in]{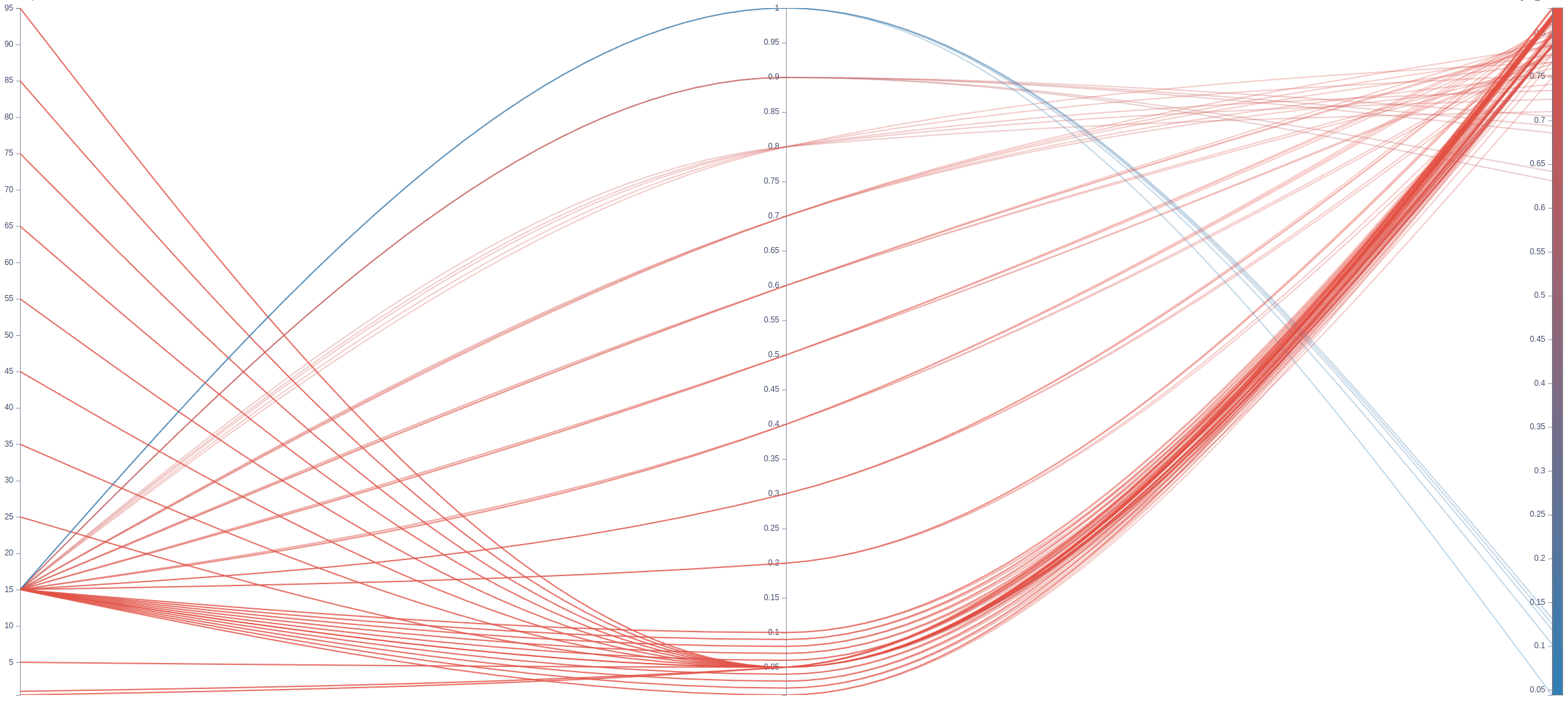}
    \includegraphics[width=0.24\textwidth,height=0.8in]{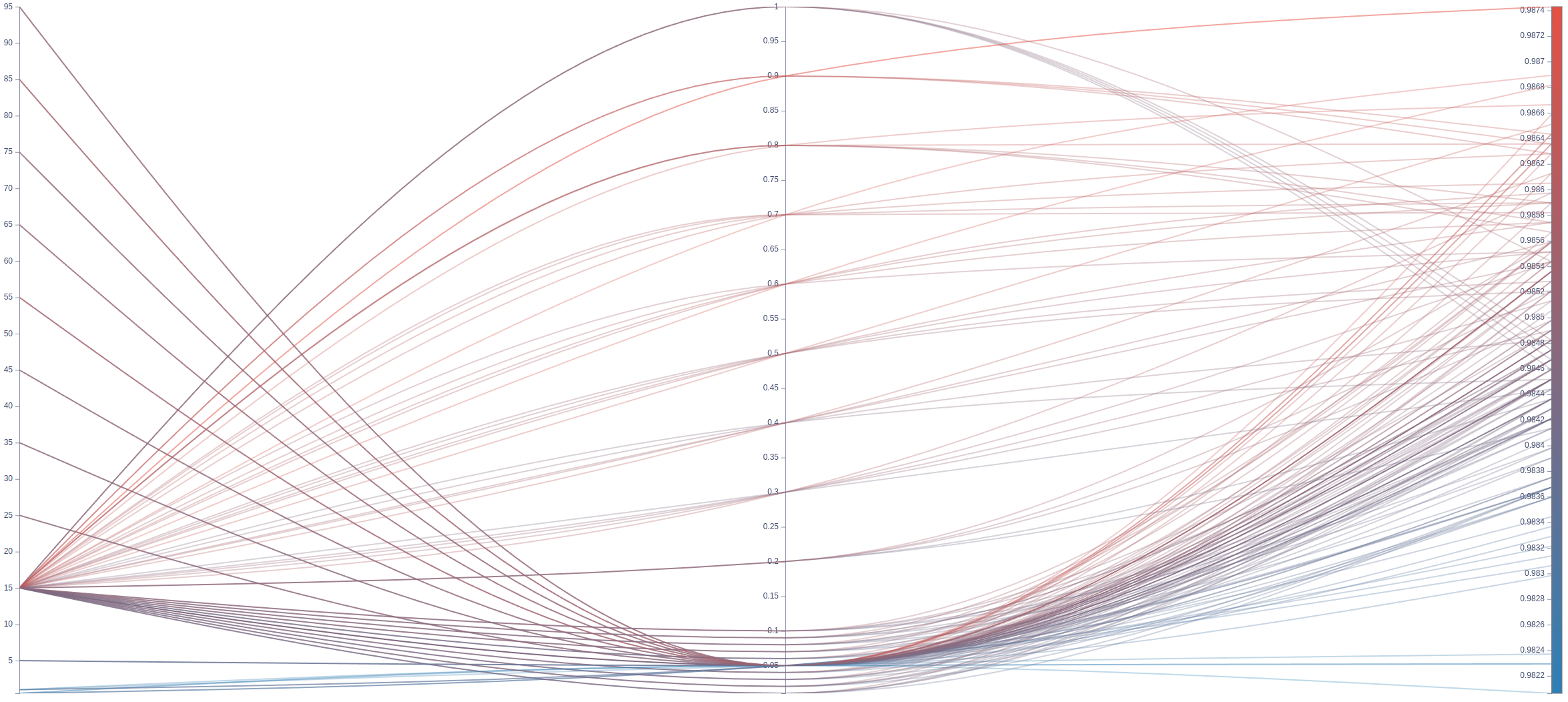}
    \includegraphics[width=0.24\textwidth,height=0.8in]{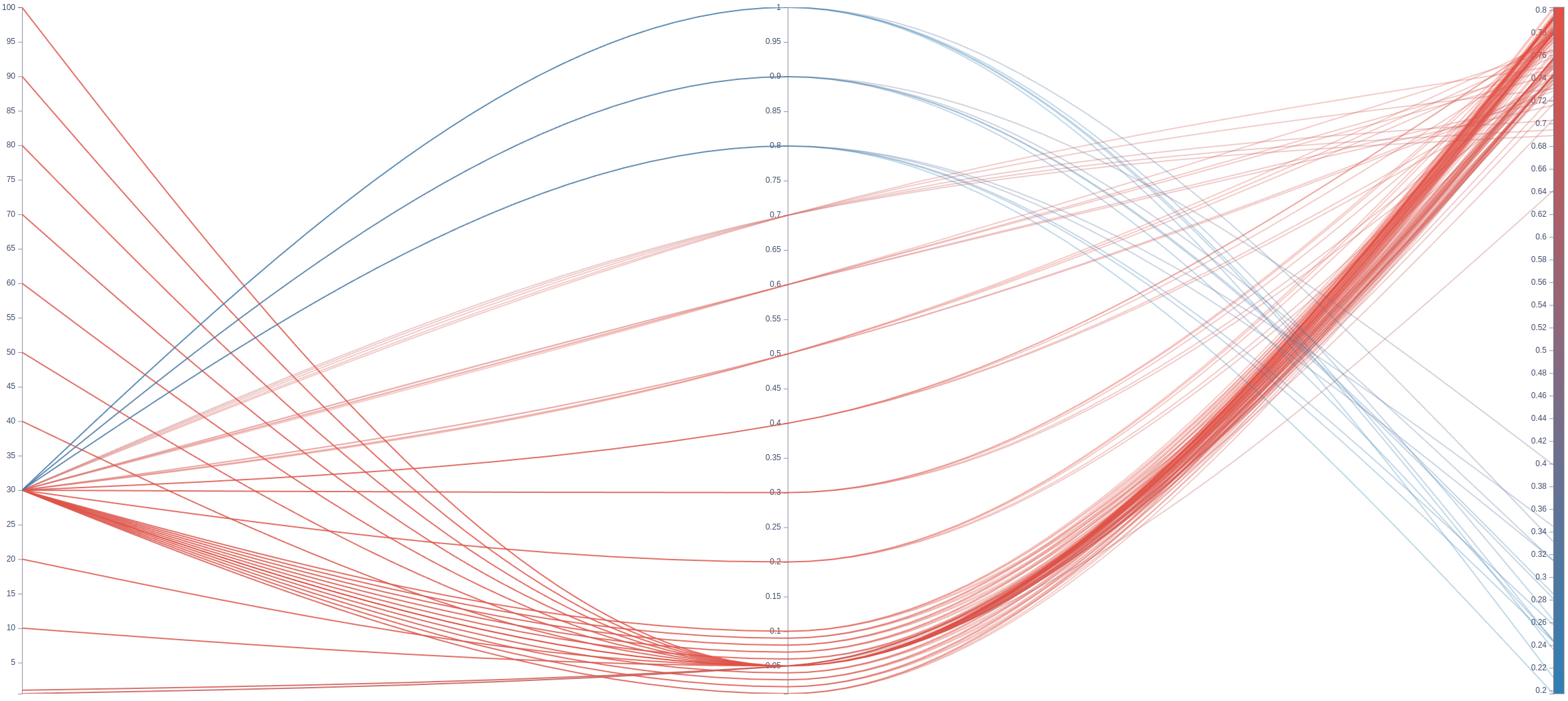}
    \includegraphics[width=0.24\textwidth,height=0.8in]{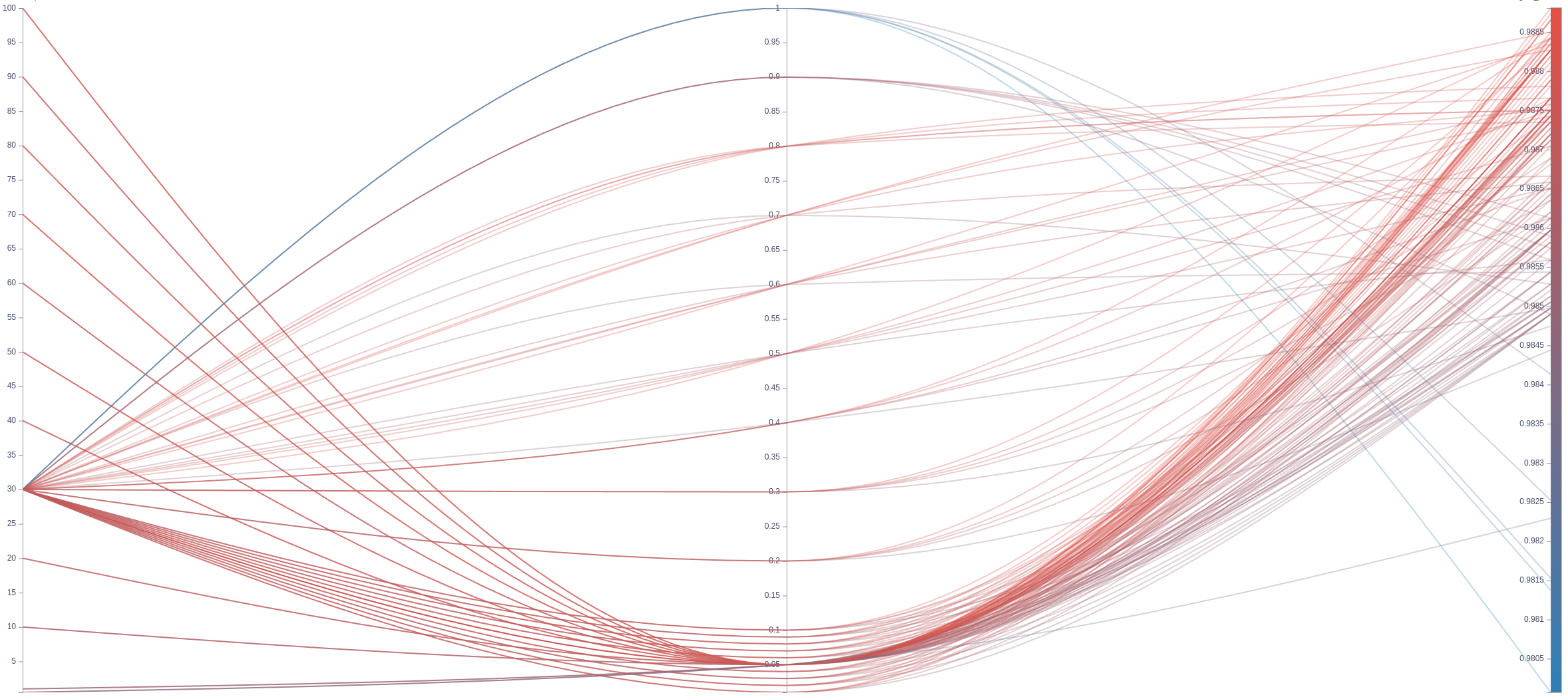}
    \begin{center}
        \footnotesize 
        FMNIST-MCLR-G \qquad\qquad\qquad FMNIST-MCLR-P
        \qquad\qquad\qquad\qquad
        FMNIST-DNN-G \qquad\qquad\qquad FMNIST-DNN-P
    \end{center}
    \caption{The left, middle and right bars in each figure respectively represent $\lambda$, $\eta$ and test accuracy, ranges of which are respectively [0,100], [0,1] and [0,1] increasing from bottom to top (color from blue to red).}
    \label{appdx_fig_hpe_fmnist}
\end{figure*}

\begin{figure*}[ht]%
    \centering
    \includegraphics[width=0.24\textwidth,height=0.8in]{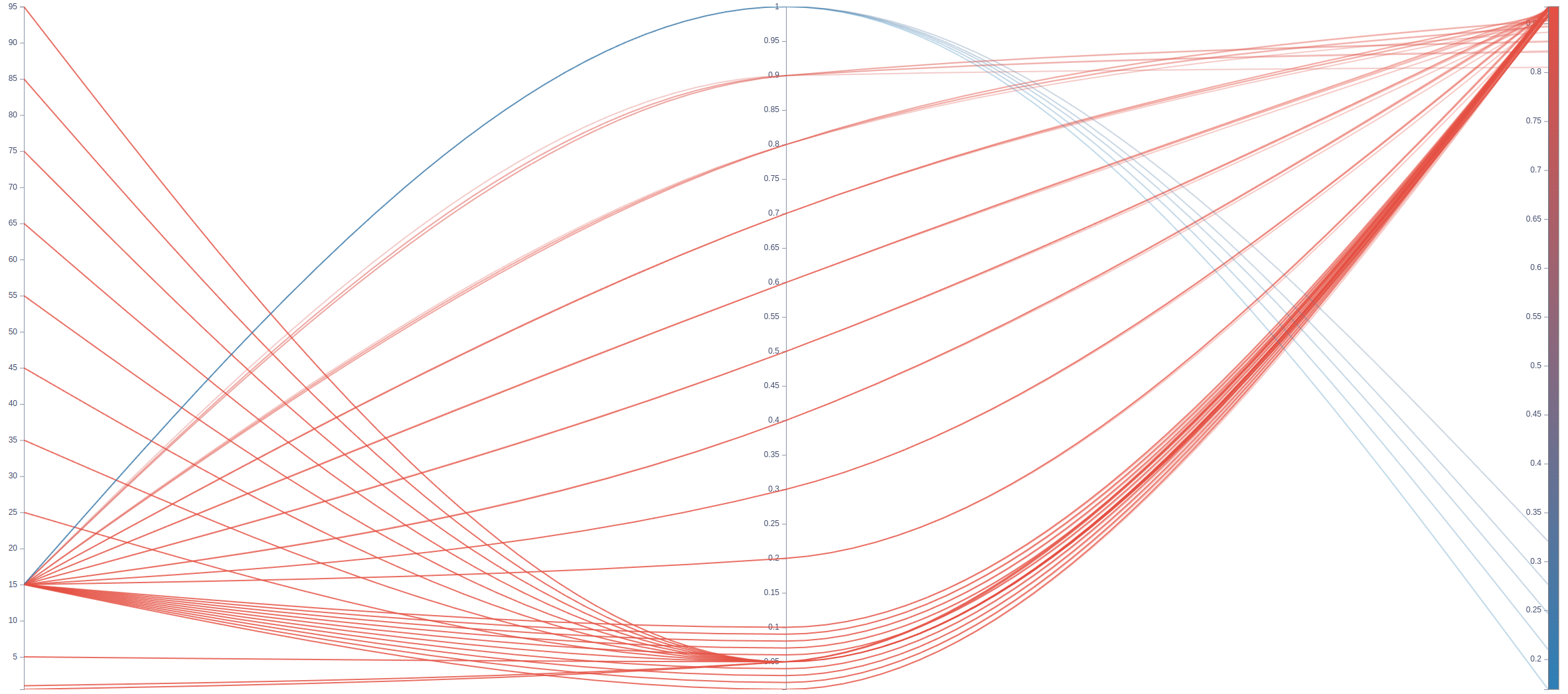}
    \includegraphics[width=0.24\textwidth,height=0.8in]{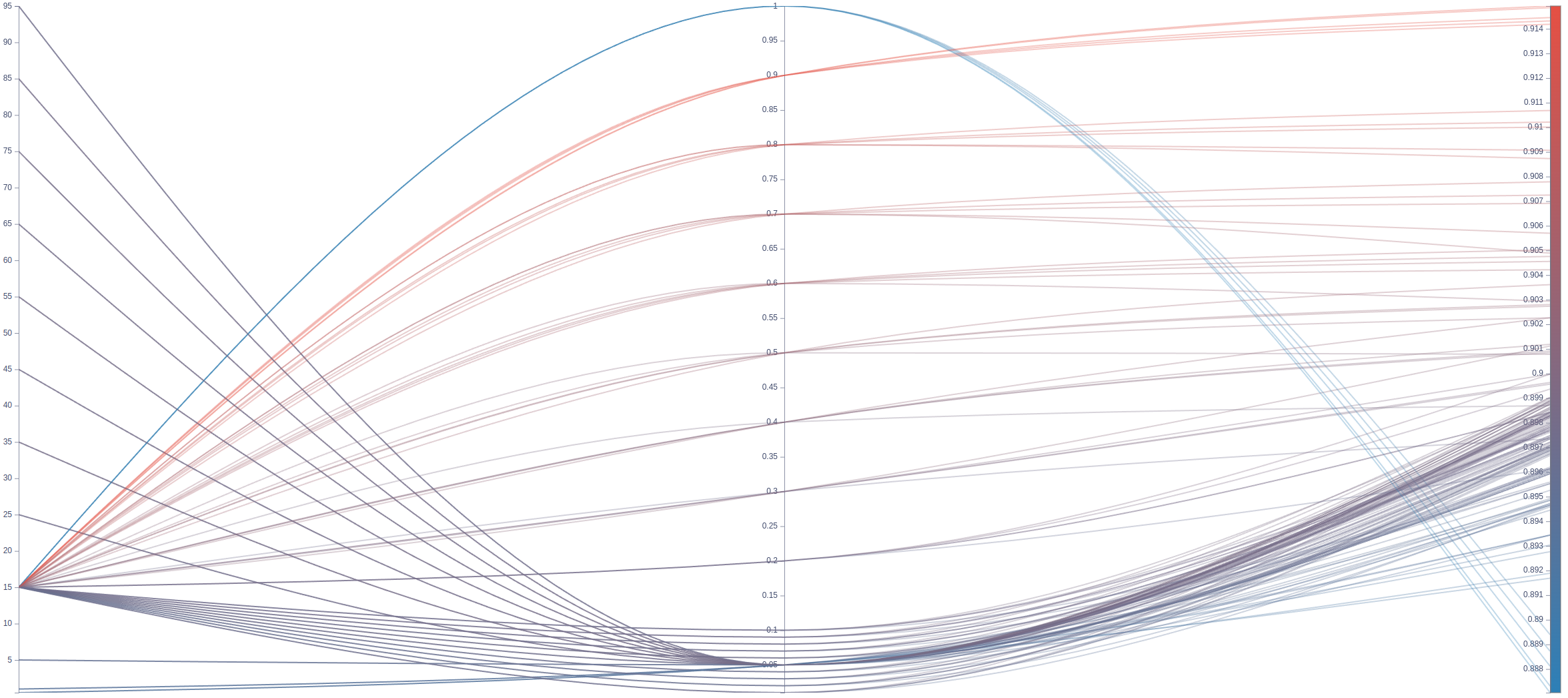}
    \includegraphics[width=0.24\textwidth,height=0.8in]{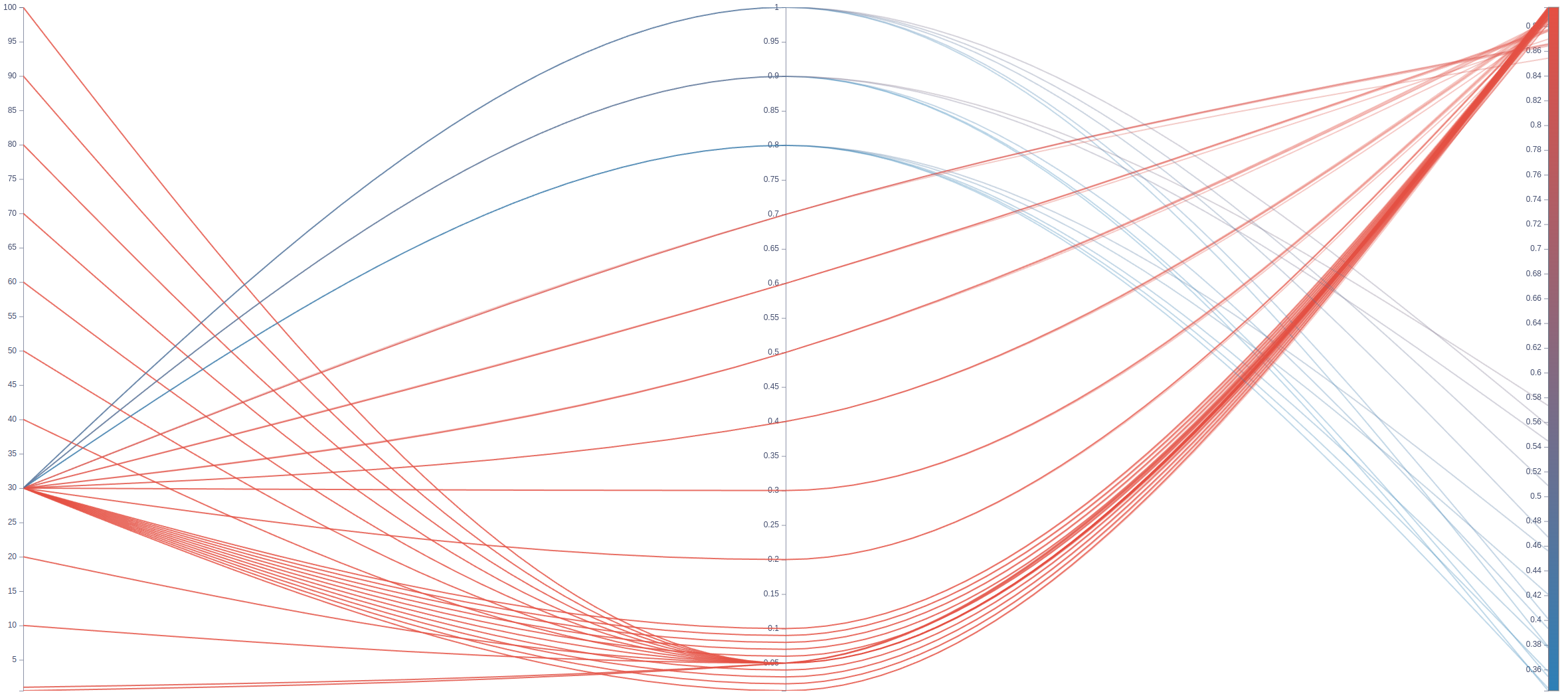}
    \includegraphics[width=0.24\textwidth,height=0.8in]{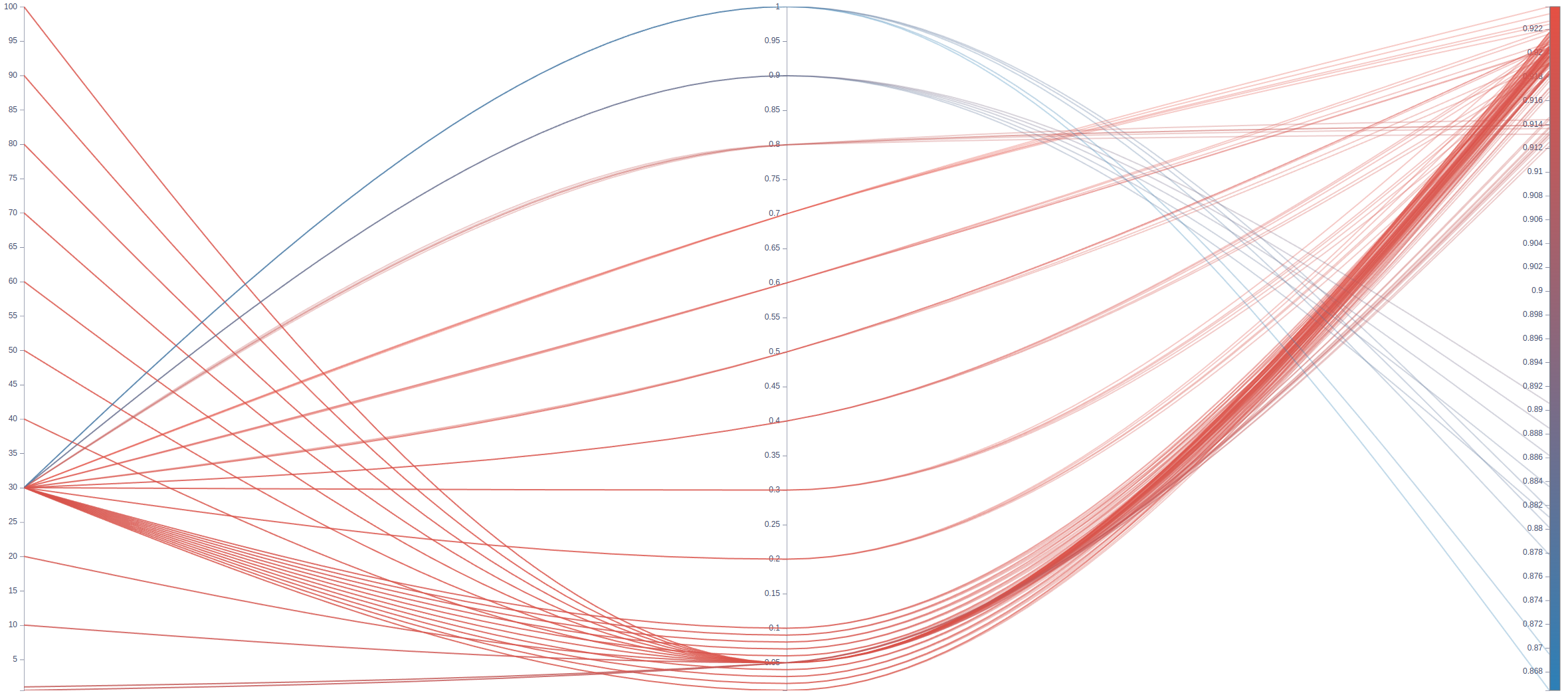}
    \begin{center}
        \footnotesize 
        MNIST-MCLR-G \qquad\qquad\qquad MNIST-MCLR-P
        \qquad\qquad\qquad\qquad
        MNIST-DNN-G \qquad\qquad\qquad MNIST-DNN-P
    \end{center}
    \caption{The left, middle and right bars in each figure respectively represent $\lambda$, $\eta$ and test accuracy, ranges of which are respectively [0,100], [0,1] and [0,1] increasing from bottom to top (color from blue to red).}
    \label{appdx_fig_hpe_mnist}
\end{figure*}

\begin{figure*}[ht]%
    \centering
    \includegraphics[width=0.24\textwidth,height=0.8in]{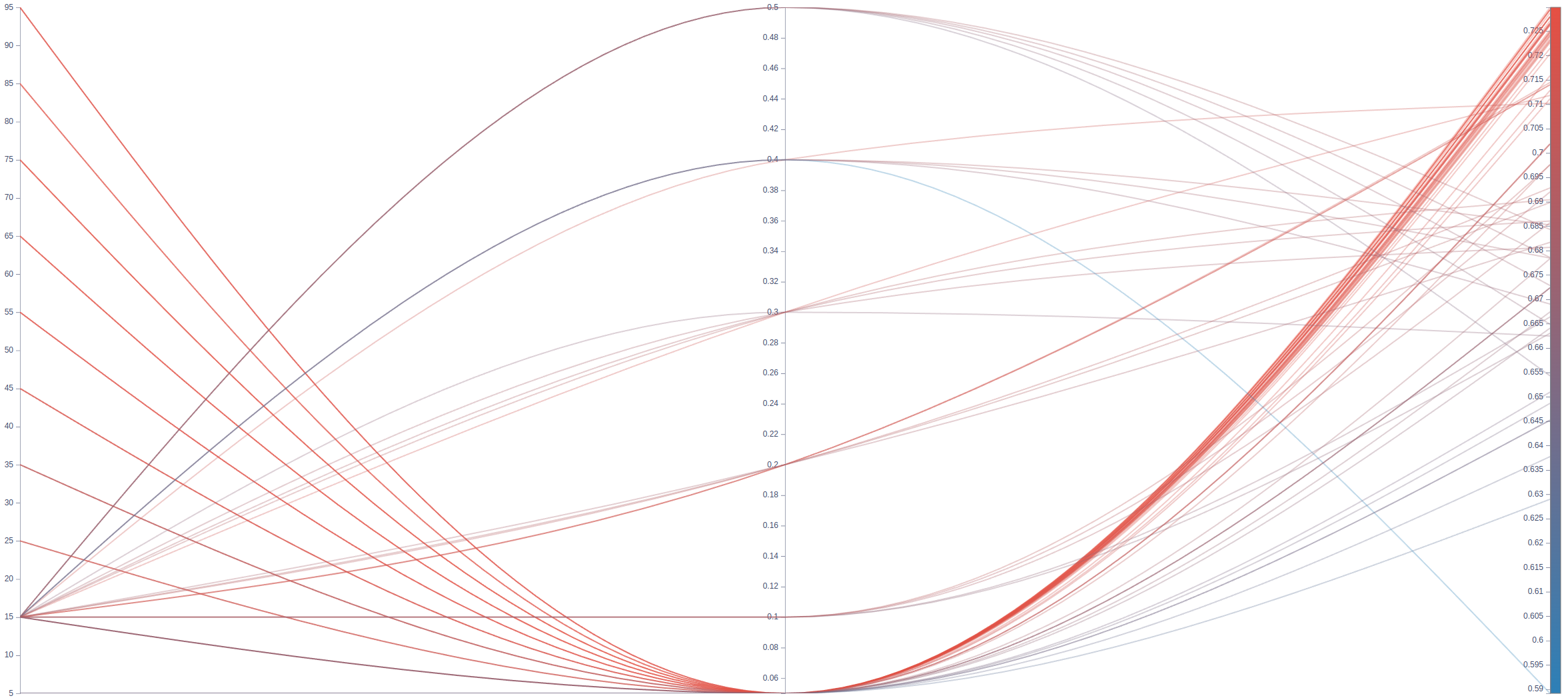}
    \includegraphics[width=0.24\textwidth,height=0.8in]{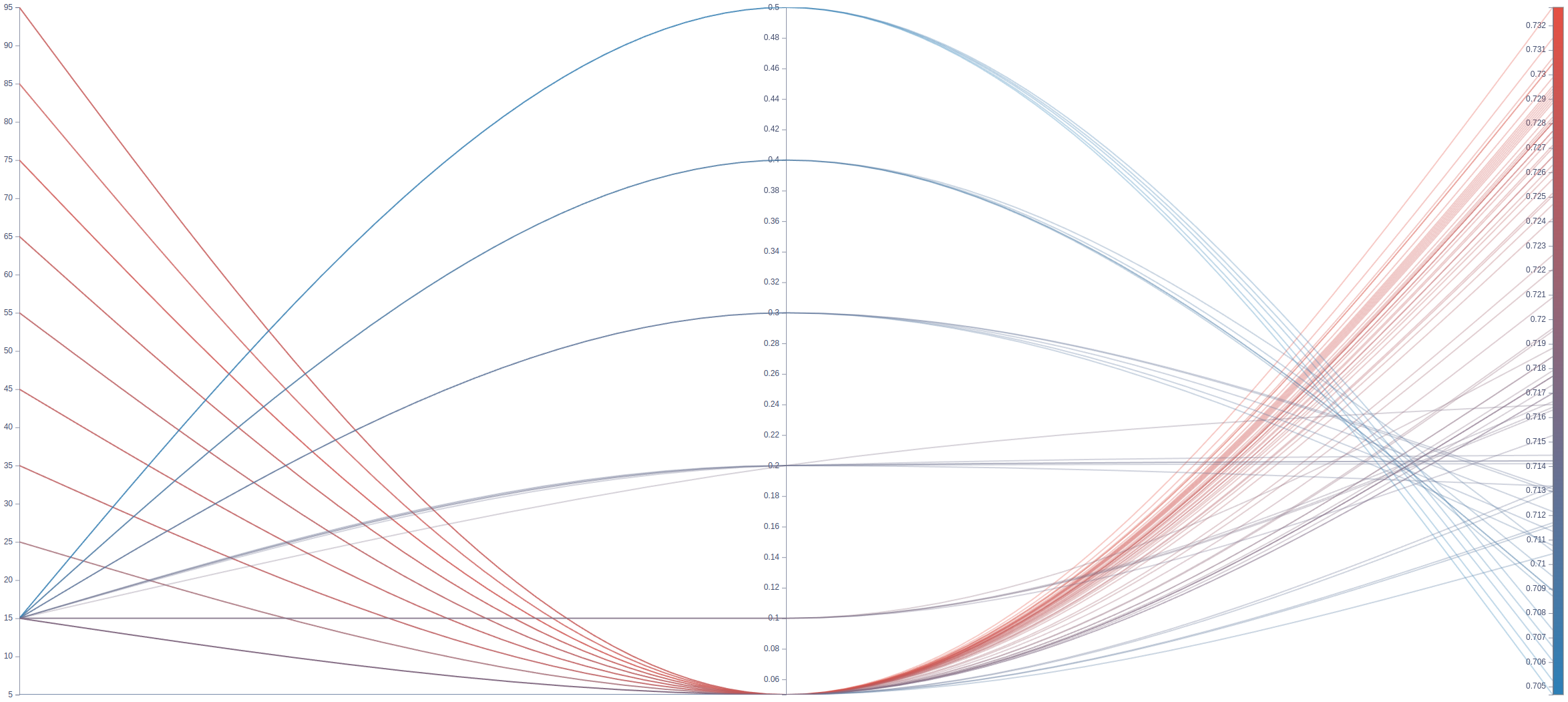}
    \includegraphics[width=0.24\textwidth,height=0.8in]{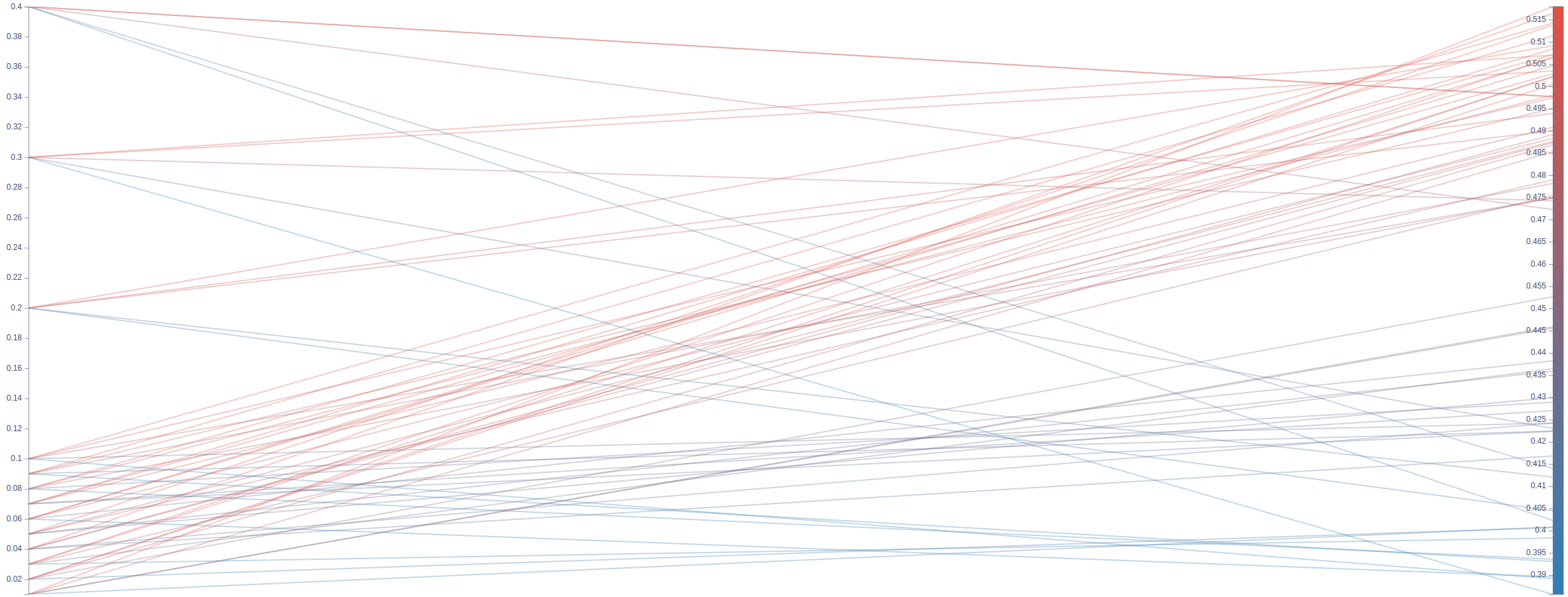}
    \includegraphics[width=0.24\textwidth,height=0.8in]{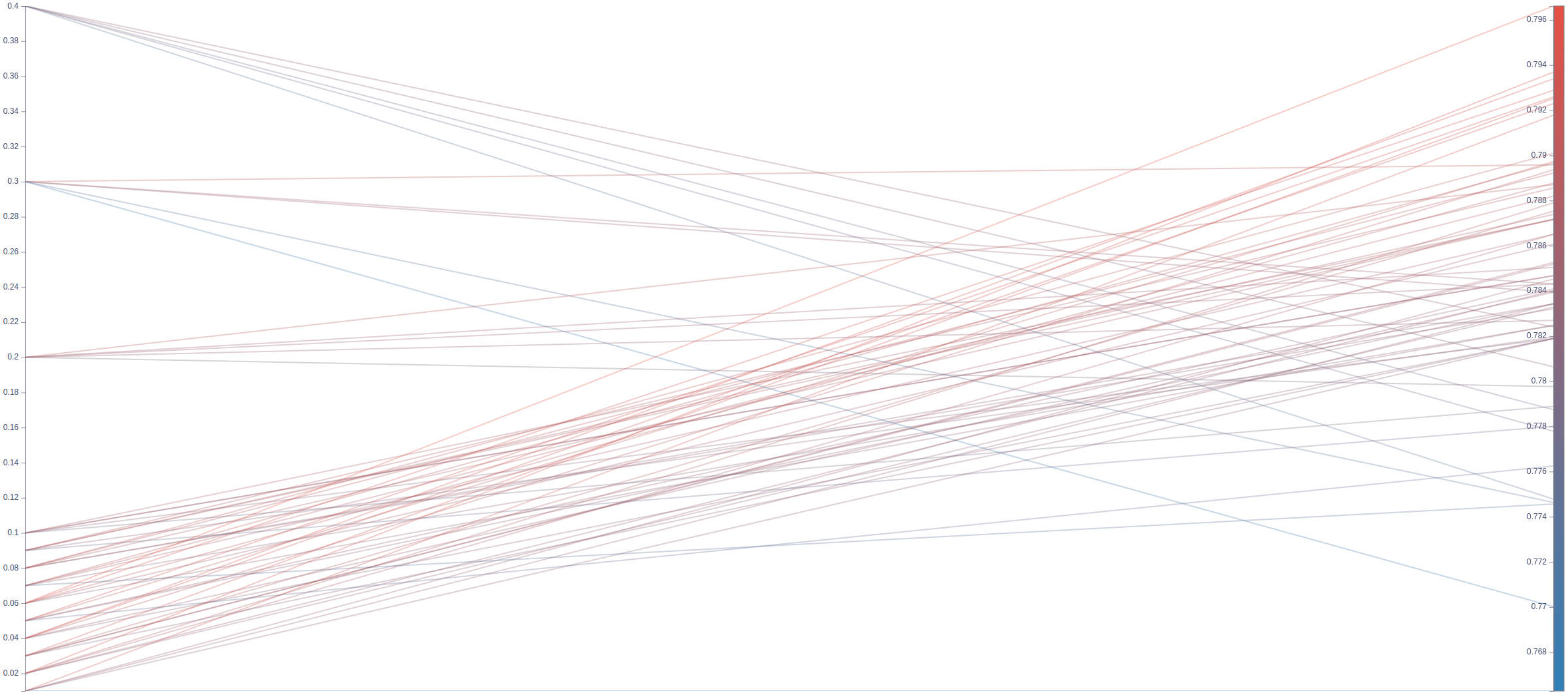}
    \begin{center}
        \footnotesize 
        Sent140-LSTM-G \qquad\qquad\qquad Sent140-LSTM-P
        \qquad\qquad\qquad\qquad
        CIFAR-10-CNN-G \qquad\qquad\qquad CIFAR-10-CNN-G
    \end{center}
    \caption{The left, middle and right bars in each figure respectively represent $\lambda$ and test accuracy, ranges of which are respectively [0,100] and [0,1] increasing from bottom to top (color from blue to red). The ranges of $\eta$ are respectively [0,0.5] and [0,0.4] in settings of CIFAR-10-CNN and Sent140.}
    \label{appdx_fig_hpe_sent140_cifar-10}
\end{figure*}

We post the hyper-parameter effects of $\eta$ and $\lambda$ on FEMNIST, FMNIST, MNIST and Sent140 and of $\eta$ on CIFAR-10 in Figure \ref{appdx_fig_hpe_femnist}-\ref{appdx_fig_hpe_sent140_cifar-10}. We haven't put the effects of $\lambda$ on CIFAR-10 for better visualization of the effects of more sensitive eta, as well as our equipment limitations, and the fact that other non-linear models for image classification are already demonstrated on FEMNIST, FMNIST and MNIST. The results of these figures are in the same hyper-parameter settings as mentioned in Section \ref{ssec_exps} except the varying hyper-parameters.

\subsection{Variant of \textbf{mg}}
\label{appdx_ssec_map}

Based on the facts, the results in Table \ref{tbl_res_lnlm} shows the instability of our personalized models. Here we propose a variant of $\mathbf{mg}$, shown in Equation (\ref{appdx_equ_v_mg}), trying to improve the robustness of personalized model on the original $\mathbf{mg}$, which use $\Phi_{i}\leftarrow f_{i}+F_{i}$.
\begin{equation}
\label{appdx_equ_v_mg}
\begin{aligned}
    \Phi_{i} & \leftarrow \tilde{F}_{i,\tilde{\eta}_{\alpha},\tilde{\eta}}:=\tilde{\eta}_{\alpha}f_{i}\circ(\cdot-\tilde{\eta}\nabla f_{i})+F_{i} \\
    \mu_{i,r} & \leftarrow w_{i,r-1}^{(t)} - \eta \nabla \tilde{F}_{i,\tilde{\eta}_{\alpha},\tilde{\eta}}(w_{i,r-1}^{(t)}) \\
    & = w_{i,r-1}^{(t)} - \eta\{ \tilde{\eta}_{\alpha} \nabla f_{i}[w_{i,r-1}^{(t)} - \tilde{\eta} \nabla f_{i}(w_{i,r-1}^{(t)})]\} - \eta \{w_{i,R}^{(t-1)}-\theta_{i,r-1}^{(t)}\}\\
\end{aligned}
\end{equation}

This method in Equation (\ref{appdx_equ_v_mg}) performance almost the same like the orginal $\mathbf{mg}$ when $\eta_{\alpha}$ is small, but it provides flexibility to tune the hyper-parameter and decide whether to focus more with the current gradient step or the meta-gradient step by tuning $\tilde{\eta}_{\alpha}$ and $\tilde{\eta}$. (We use $\tilde{\eta}_{\alpha}\leftarrow \eta_{\alpha}/\eta$ and $\tilde{\eta}\leftarrow \eta_{\alpha}$ in practice, which are not well-tune.)

\begin{figure*}[ht]
    \centering
    \includegraphics[width=0.24\textwidth,height=0.8in]{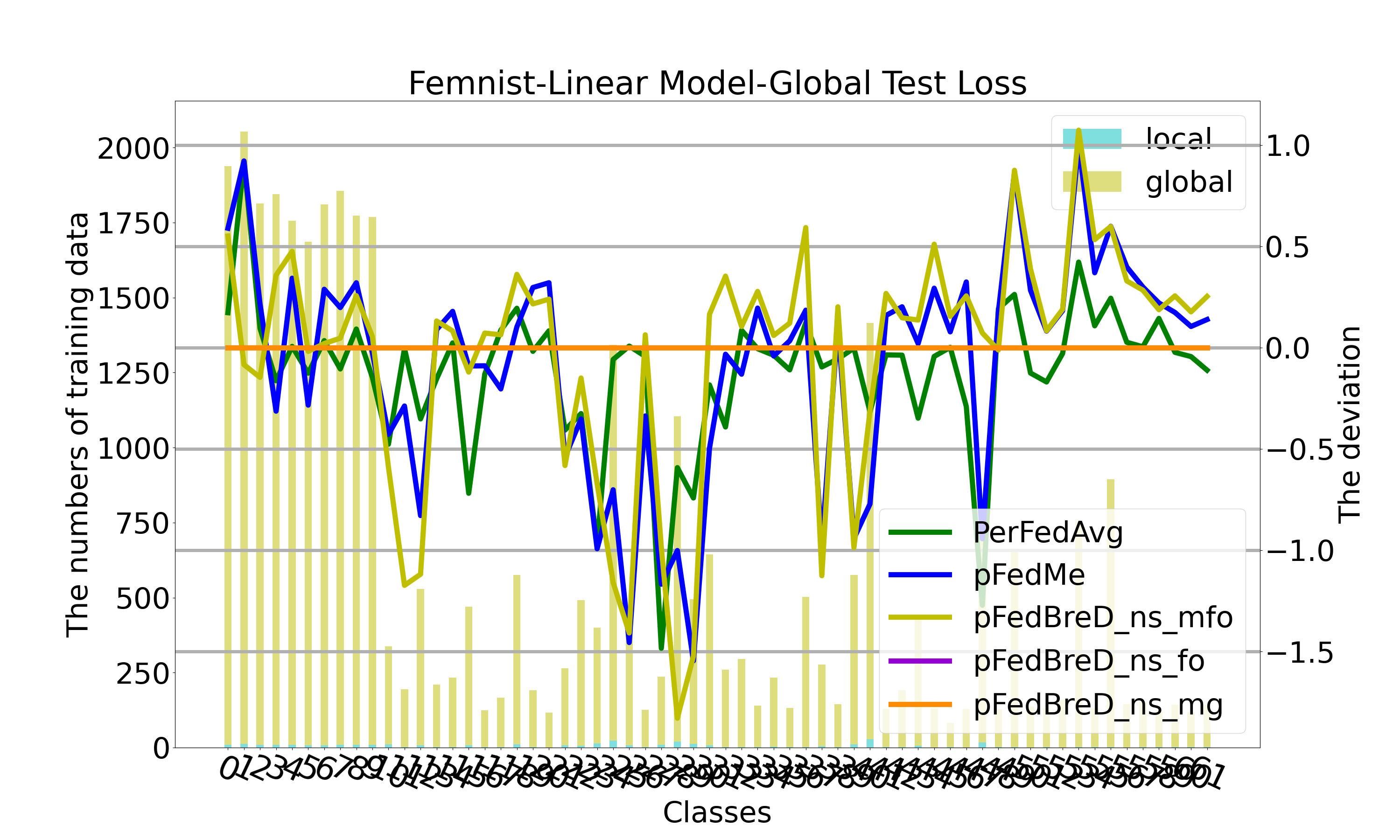}
    \includegraphics[width=0.24\textwidth,height=0.8in]{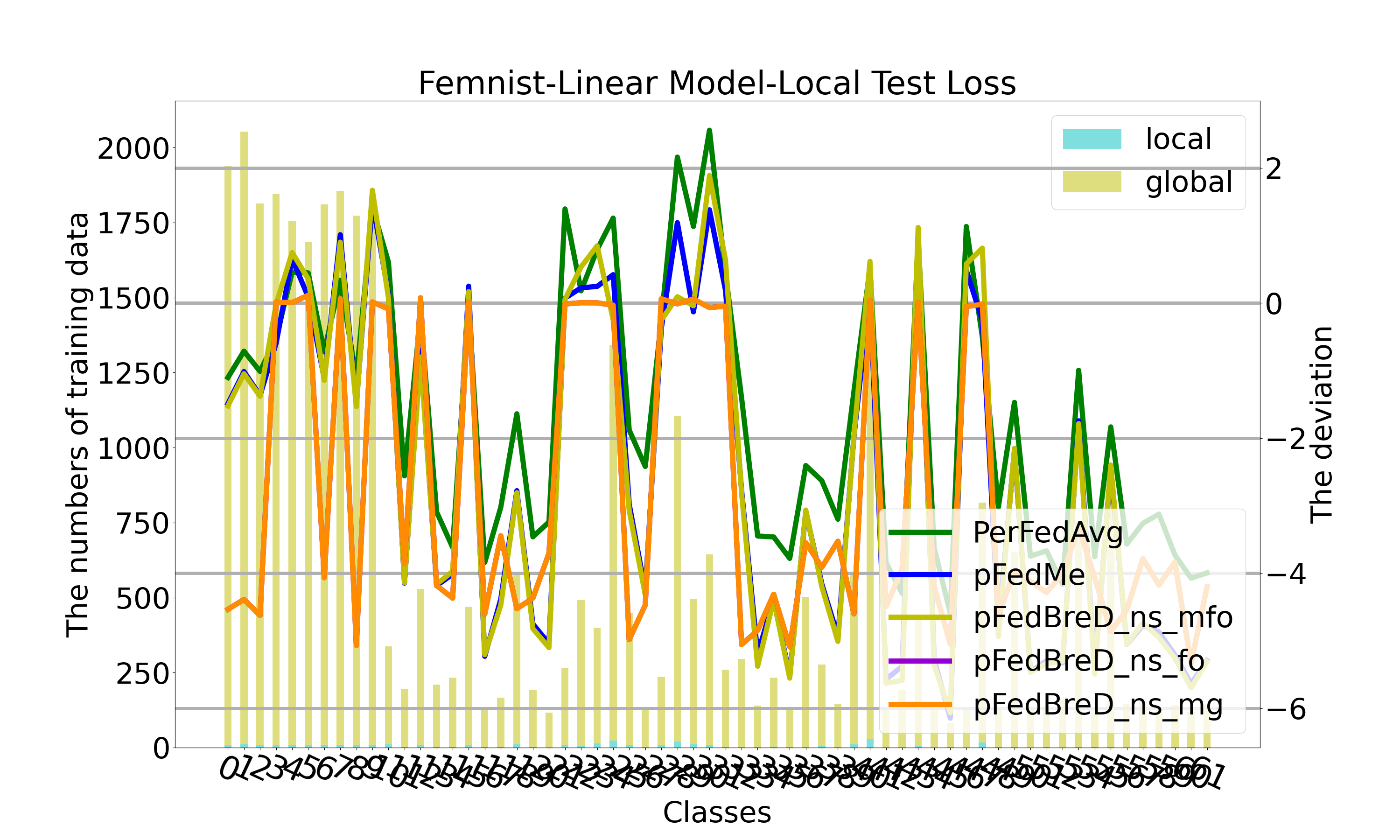}
    \includegraphics[width=0.24\textwidth,height=0.8in]{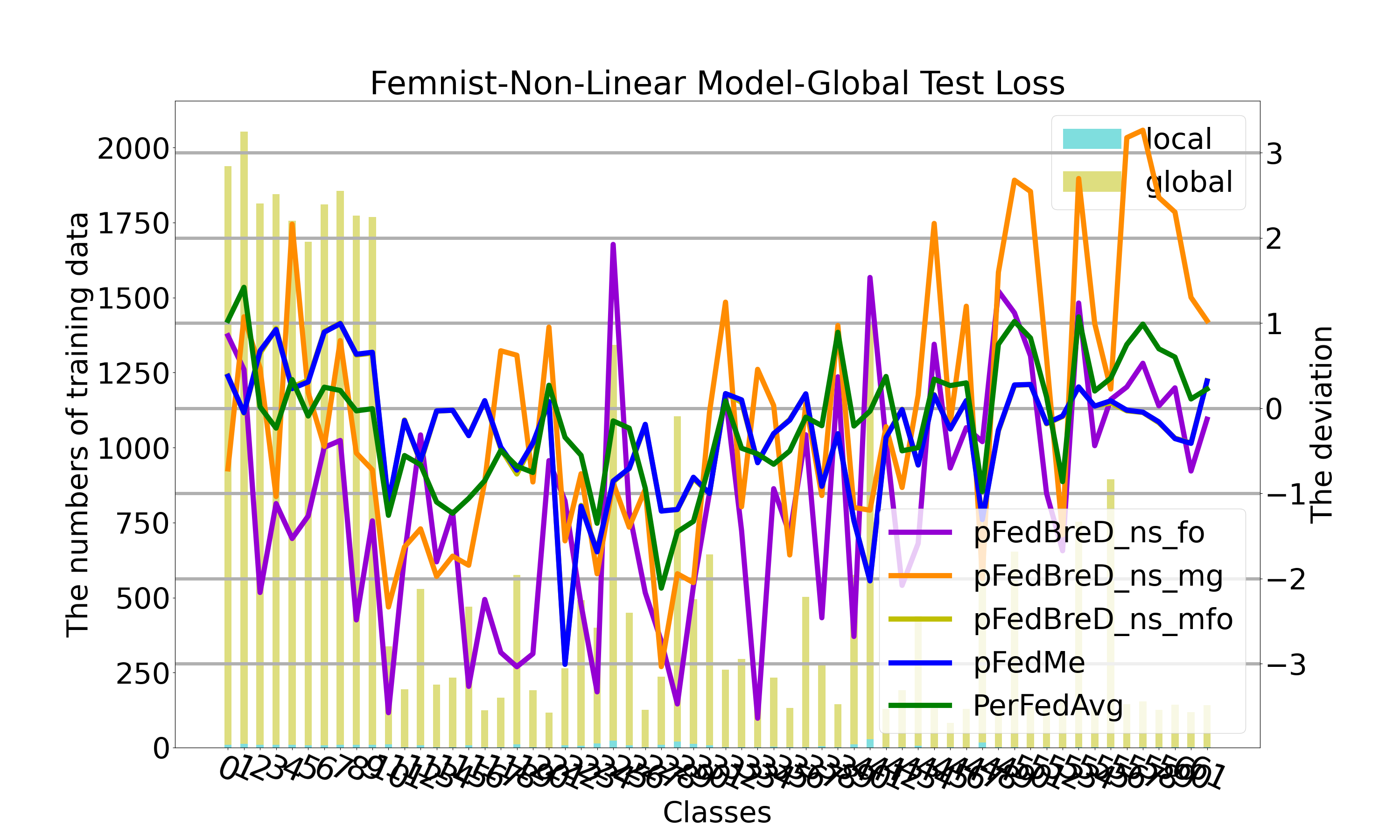}
    \includegraphics[width=0.24\textwidth,height=0.8in]{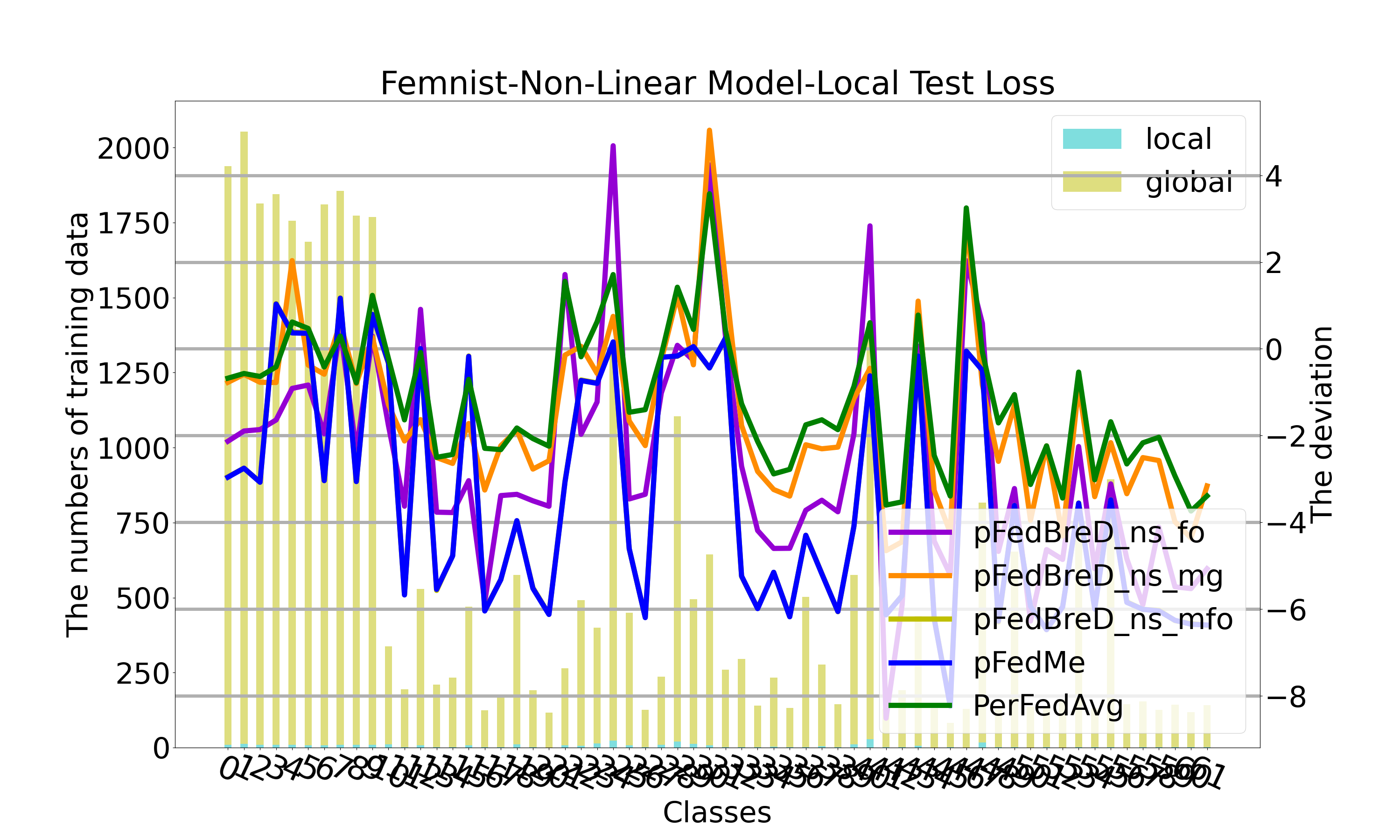}
    \begin{center}
        \footnotesize 
        FEMNIST-MCLR-G \qquad\qquad\qquad FEMNIST-MCLR-L
        \qquad\qquad\qquad\qquad
        FEMNIST-DNN-G \qquad\qquad\qquad FEMNIST-DNN-L
    \end{center}
    \includegraphics[width=0.24\textwidth,height=0.8in]{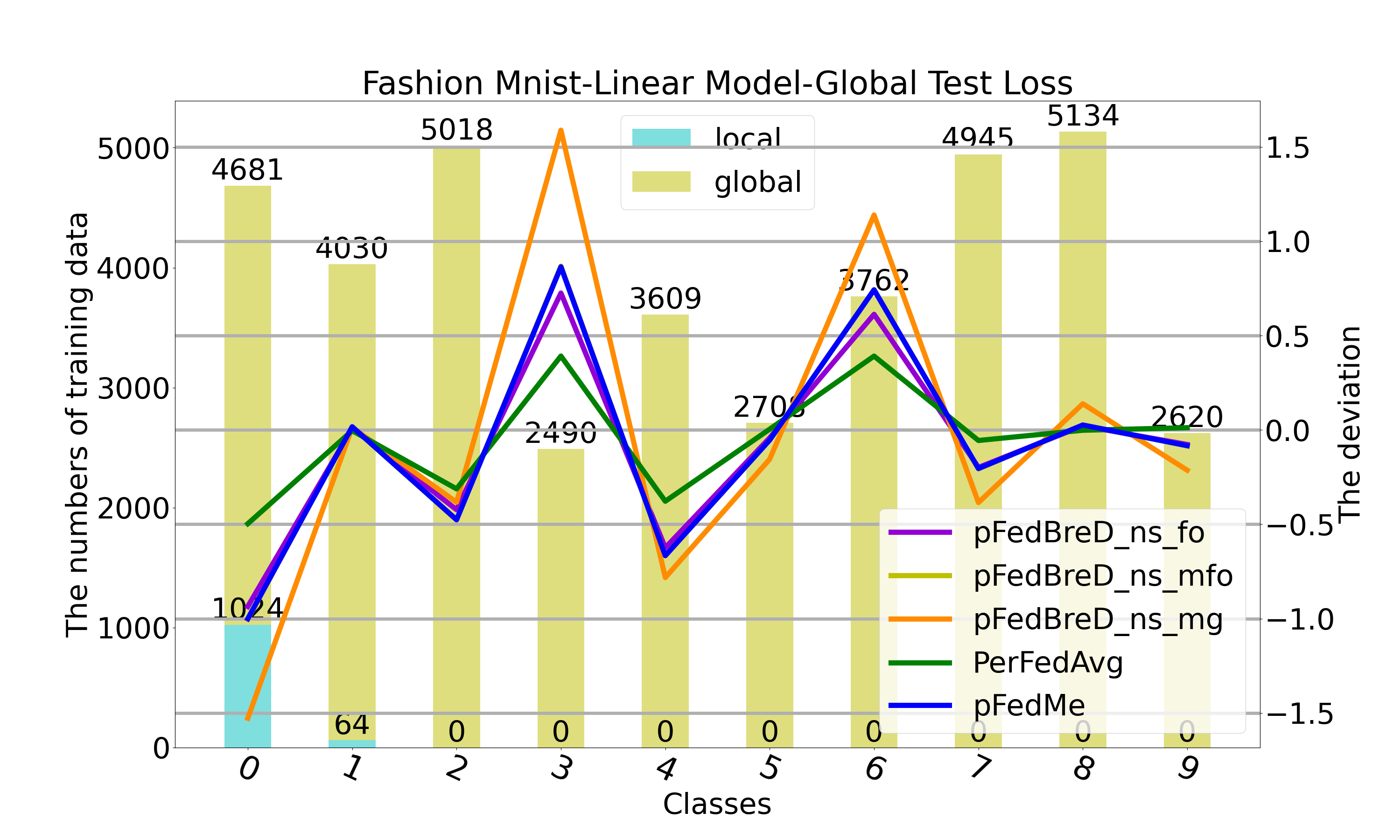}
    \includegraphics[width=0.24\textwidth,height=0.8in]{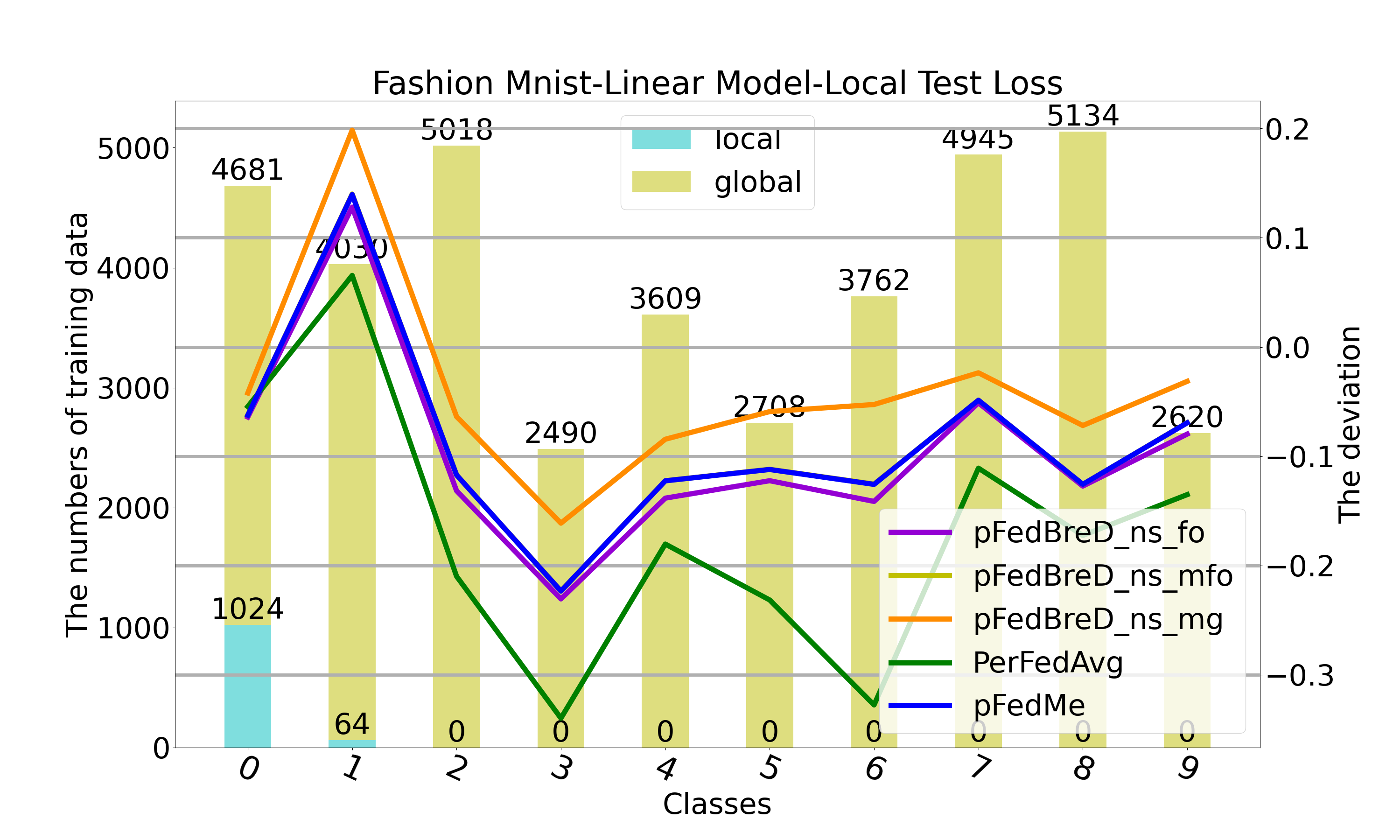}
    \includegraphics[width=0.24\textwidth,height=0.8in]{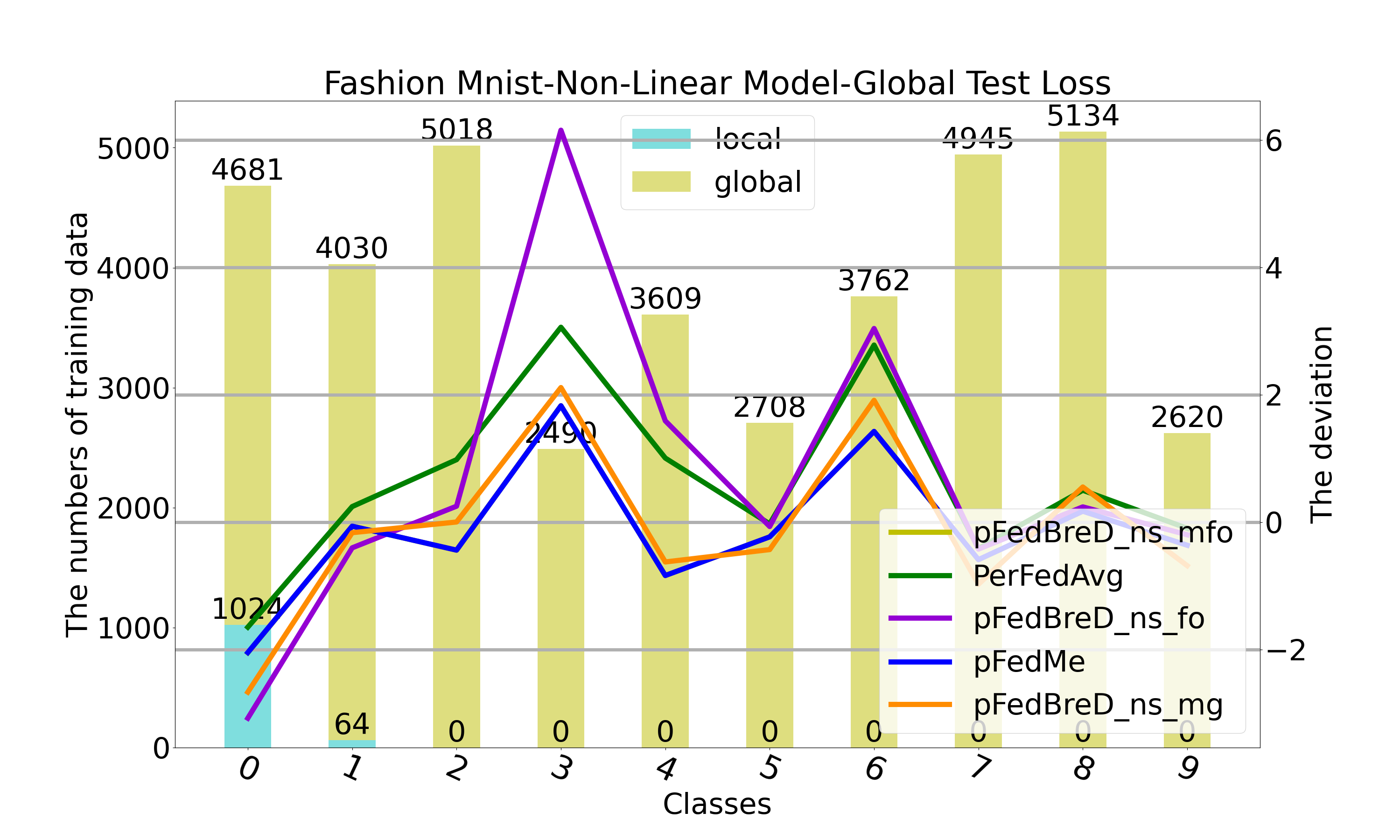}
    \includegraphics[width=0.24\textwidth,height=0.8in]{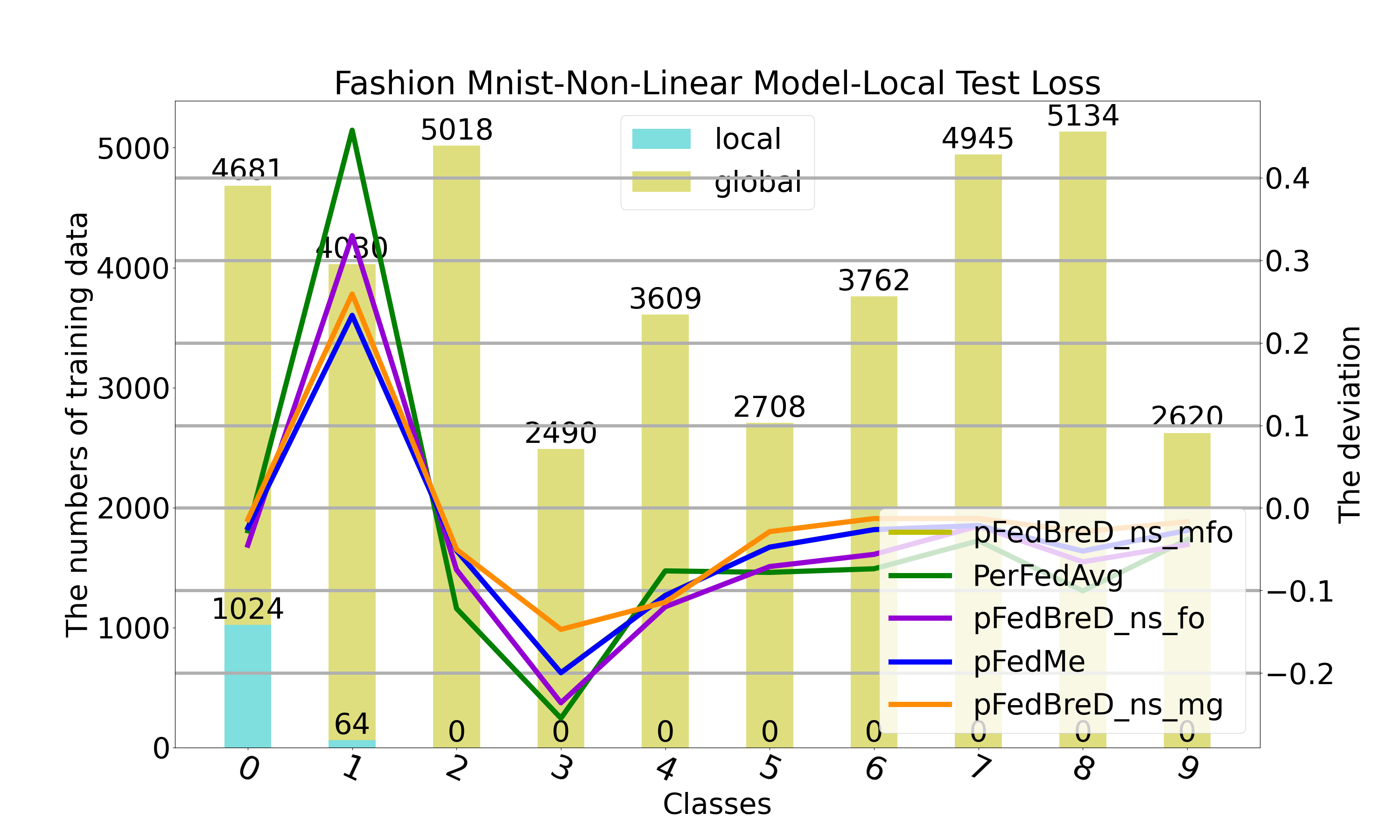}
    \begin{center}
        \footnotesize 
        FMNIST-MCLR-G \qquad\qquad\qquad FMNIST-MCLR-L
        \qquad\qquad\qquad\qquad
        FMNIST-DNN-G \qquad\qquad\qquad FMNIST-DNN-L
    \end{center}
    \includegraphics[width=0.24\textwidth,height=0.8in]{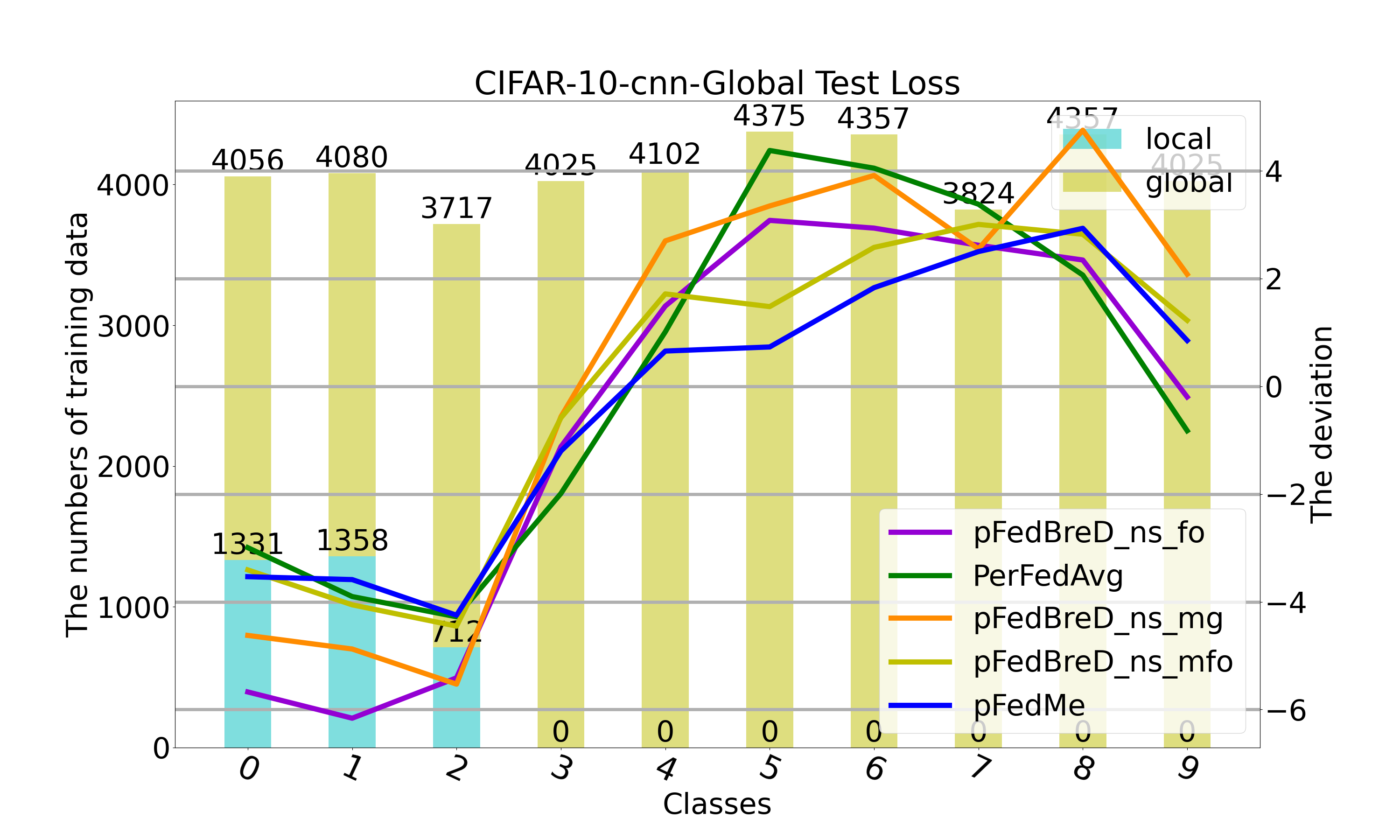}
    \includegraphics[width=0.24\textwidth,height=0.8in]{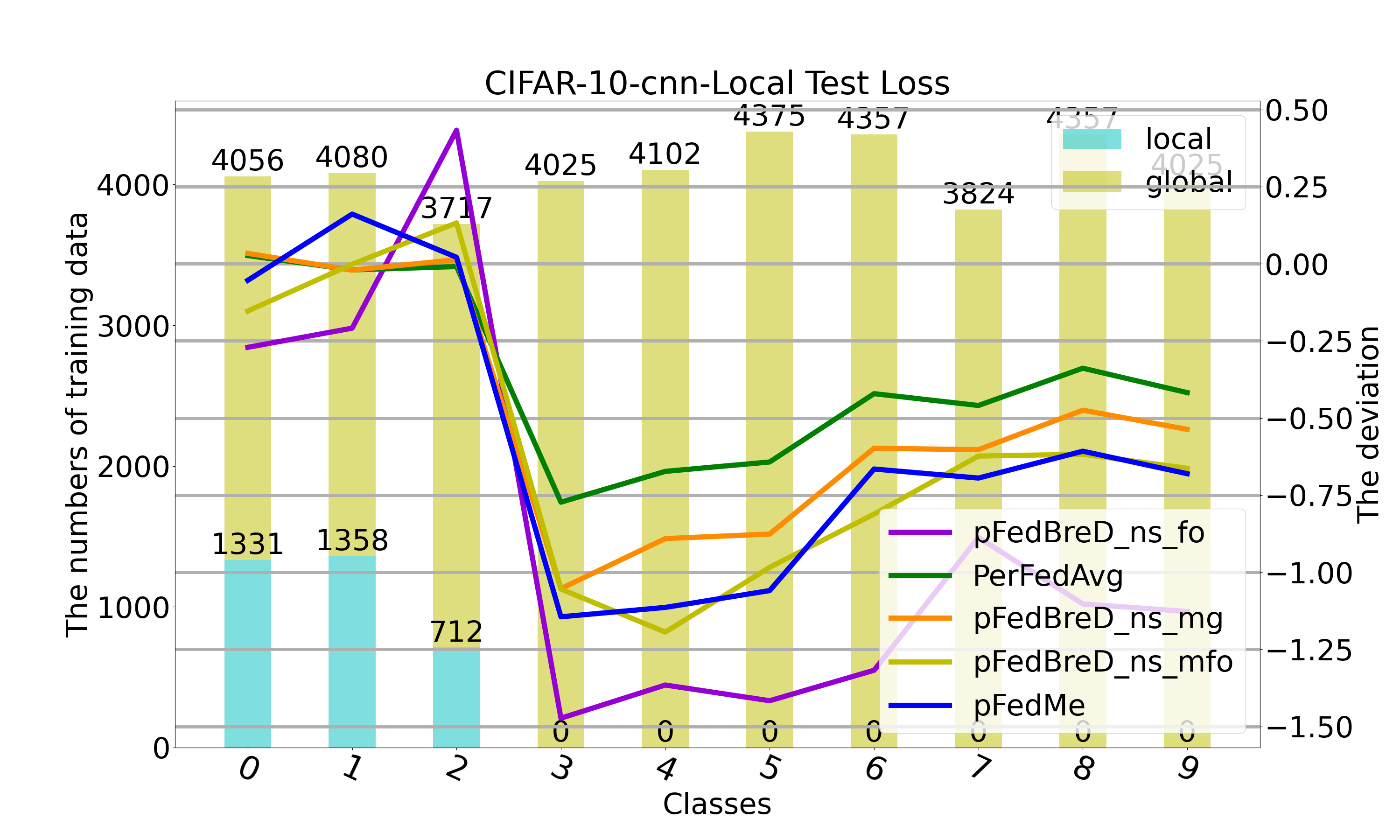}
    \includegraphics[width=0.24\textwidth,height=0.8in]{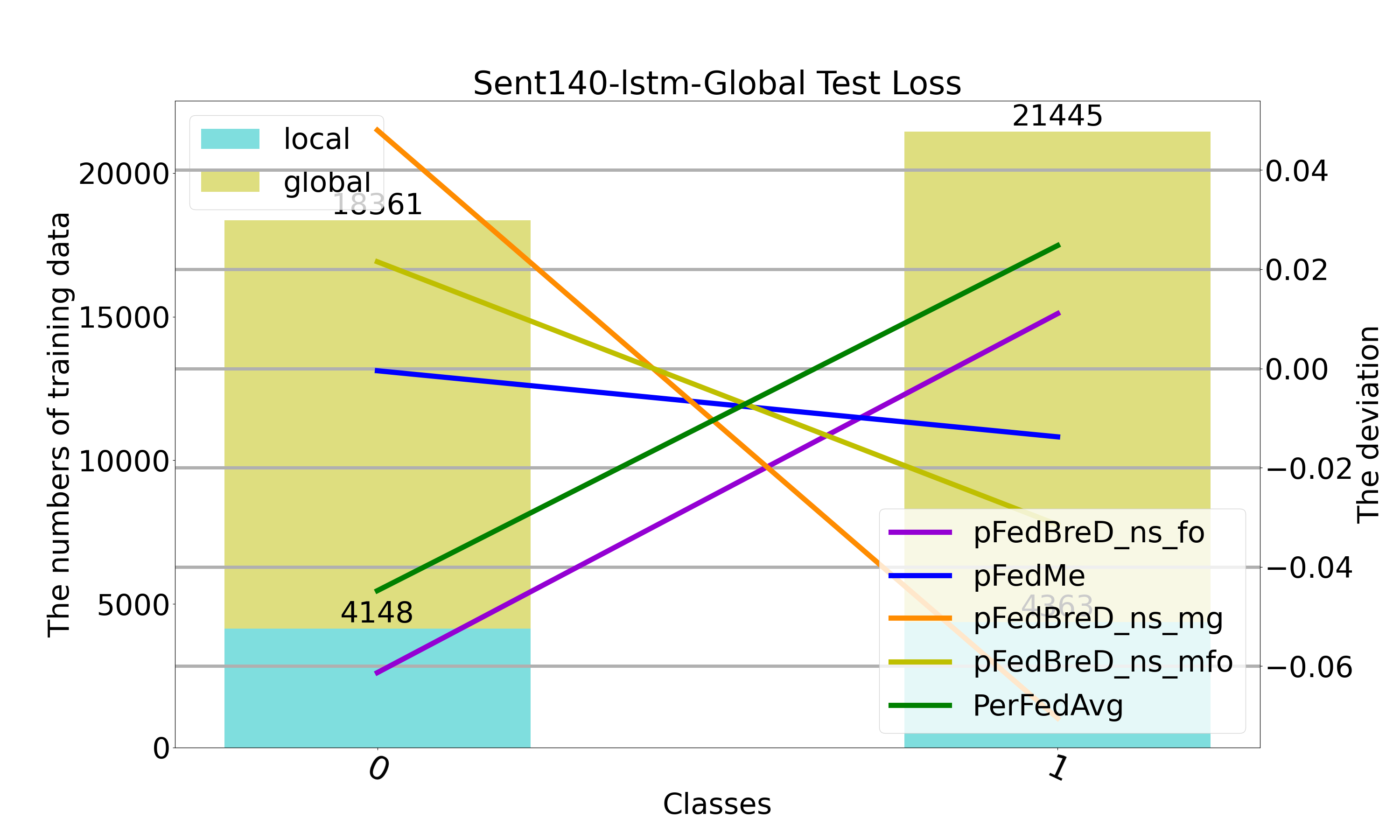}
    \includegraphics[width=0.24\textwidth,height=0.8in]{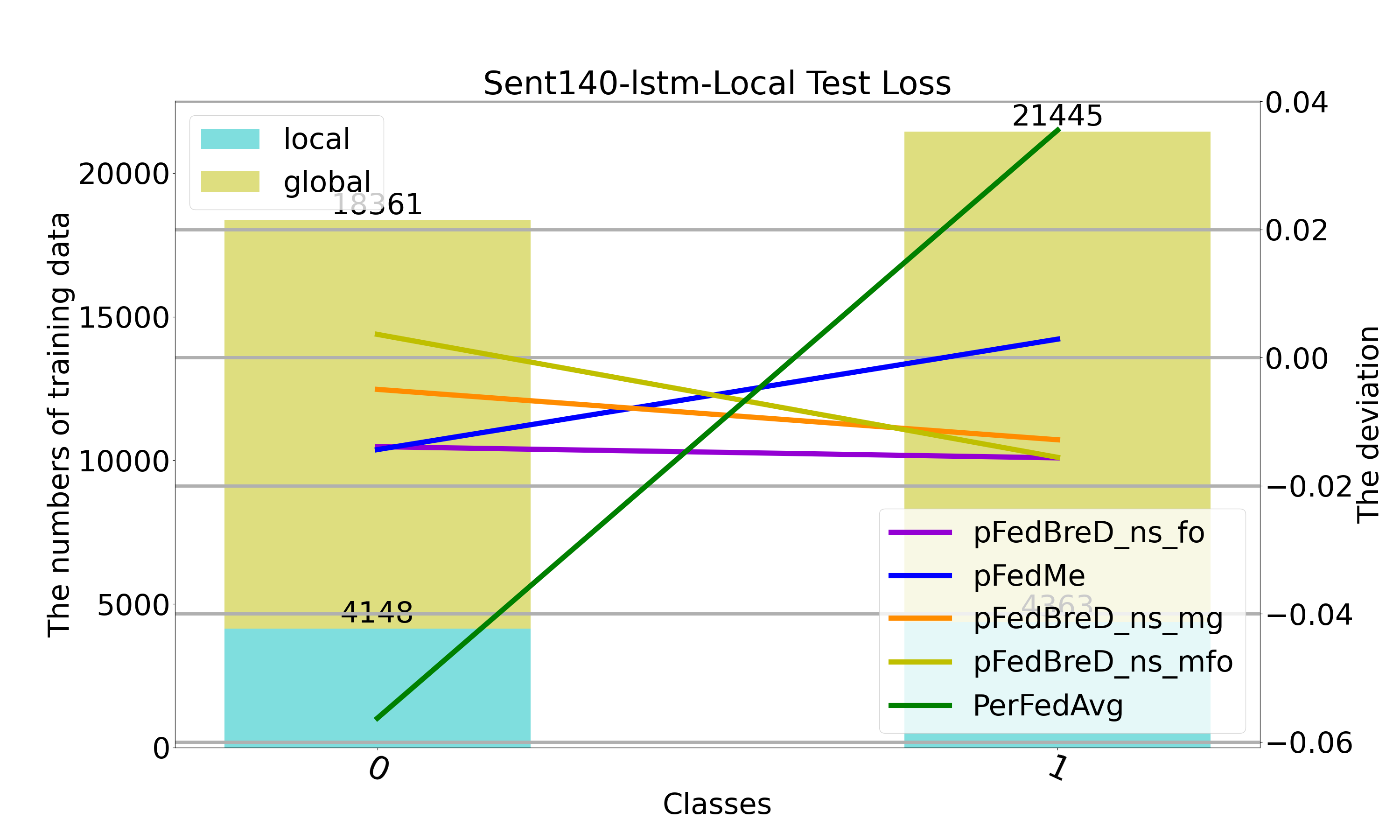}
    \begin{center}
        \footnotesize 
        CIFAR-10-CNN-G \qquad\qquad\qquad CIFAR-10-CNN-L
        \qquad\qquad\qquad\qquad
        Sent140-LSTM-G \qquad\qquad\qquad Sent140-LSTM-L
    \end{center}
    \caption{The loss deviation of our experiments in Section \ref{ssec_expr} on the first client on settings: FEMNIST-DNN/MCLR, FMNIST-DNN/MCLR, CIFAR-10-CNN and Sent140-LSTM.}
    \label{appdx_fig_psnlz_com}
\end{figure*}

The deviations of the global and local test on each settings are shown in Figure \ref{appdx_fig_psnlz_com} mentioned in Section \ref{ssec_personalization} in the main paper.

\subsection{Instability}
\label{appdx_ssec_instability}

This section we experimentally show the instability of the global model in $\mathbf{mg}$ on small aggregation ratio by compare the performances on different aggregation numbers of the clients at the end of each global epoch. We use the settings in Section \ref{ssec_exps} in the main paper except the numbers of clients for aggregation at the end of each global epoch, and the maximum results of the experiments, in each setting are shown in Table \ref{appdx_tbl_aggregation}, where the additional experiments were repeated twice.

\begin{table}[t]
    \centering
    \resizebox{.94\textwidth}{!}{
    \begin{tabular}{l|c|c|c|c}
        \hline
        \textbf{Numbers} & 10 & 20 & 50 & 100\\
        \hline
        FEMNIST-DNN & 58.96 & 60.08 & 60.06 & 59.82 \\
        FEMNIST-MCLR & 54.43 & 55.38 & 55.41 & 55.52 \\
        FMNIST-DNN & 75.07 & 79.55 & 79.37 & 79.34 \\
        FMNIST-MCLR & 79.96 & 82.58 & 81.77 & 82.72 \\
        \hline
    \end{tabular}
    }
    \caption{The global test accuracy (\%) of the global model with different numbers of clients for aggregation at the same not-converged global epoch.}
    \label{appdx_tbl_aggregation}
\end{table}


\end{document}